\pdfoutput=1
\documentclass[11pt]{article}

\usepackage[preprint]{acl}

\usepackage{times}
\usepackage{latexsym}

\usepackage[T1]{fontenc}
\usepackage[utf8]{inputenc}

\usepackage{microtype}

\usepackage{inconsolata}

\usepackage{graphicx}
\usepackage{microtype}
\usepackage{float}
\usepackage{array}    
\usepackage{siunitx}  
\usepackage{multirow}
\usepackage{graphicx}
\usepackage{subcaption}
\usepackage{listings}
\lstdefinelanguage{isabelle}{}
\usepackage{xcolor}
\usepackage{booktabs}
\usepackage{longtable}
\usepackage{array}
\usepackage{xcolor}
\usepackage{tcolorbox}
\usepackage{amsmath, amssymb}
\usepackage{newunicodechar}            
\usepackage{makecell}          
\usepackage{booktabs}  
\usepackage{arydshln}   
\usepackage{siunitx}

\usepackage[utf8]{inputenc}
\usepackage{qtree}
\usepackage[ruled, lined, linesnumbered, commentsnumbered, longend]{algorithm2e}

\definecolor{darkgreen}{RGB}{0,100,0}
\title{Faithful and Robust LLM-Driven Theorem Proving for NLI Explanations}

\author{Xin Quan$^1$, Marco Valentino$^{2,3}$, Louise A. Dennis$^1$, Andr\'e Freitas$^{1,2,4}$ \\ 
$^{1}$Department of Computer Science, University of Manchester, UK \\ 
$^{2}$Idiap Research Institute, Switzerland \\
$^{3}$School of Computer Science, University of Sheffield, UK \\ 
$^{4}$National Biomarker Centre, CRUK-MI, University of Manchester, UK\\
$^1$\tt{\{name.surname\}@manchester.ac.uk}\\
$^2$\tt{\{name.surname\}@idiap.ch}}

\begin{document}
\maketitle
\begin{abstract}
Natural language explanations play a fundamental role in Natural Language Inference (NLI) by revealing how premises logically entail hypotheses. Recent work has shown that the interaction of large language models (LLMs) with theorem provers (TPs) can help verify and improve the validity of NLI explanations. However, TPs require translating natural language into machine-verifiable formal representations, a process that introduces the risk of semantic information loss and unfaithful interpretation, an issue compounded by LLMs' challenges in capturing critical logical structures with sufficient precision. Moreover, LLMs are still limited in their capacity for rigorous and robust proof construction within formal verification frameworks. To mitigate issues related to faithfulness and robustness, this paper investigates strategies to (1) alleviate semantic loss during autoformalisation, (2) efficiently identify and correct syntactic errors in logical representations, (3) explicitly use logical expressions to guide LLMs in generating structured proof sketches, and (4) increase LLMs' capacity of interpreting TP's feedback for iterative refinement. Our empirical results on e-SNLI, QASC and WorldTree using different LLMs demonstrate that the proposed strategies yield significant improvements in autoformalisation (+18.46\%, +34.2\%, +39.77\%) and explanation refinement (+29.5\%, +51.5\%, +41.25\%) over the state-of-the-art model. Moreover, we show that specific interventions on the hybrid LLM-TP architecture can substantially improve efficiency, drastically reducing the number of iterations required for successful verification.\footnote{Code and data are available at: \href{https://github.com/neuro-symbolic-ai/faithful_and_robust_nli_refinement}{https://github.com/neuro\-symbolic\-ai/faithful\_and\_robust\_nli\_refinement}}
\end{abstract}

\section{Introduction}
Recent studies in Natural Language Inference (NLI) have developed models to leverage natural language explanations as a mechanism for reasoning in support of a hypothesis \citep{wiegreffe2021teachexplainreviewdatasets, chen-etal-2021-kace, thayaparan2020surveyexplainabilitymachinereading, valentino-etal-2022-case}. Providing sound and logically valid natural language explanations lies at the core of NLI, as such transparent justifications enhance both interpretability and reliability for downstream tasks \citep{NEURIPS2018_4c7a167b,valentino-etal-2022-case, he-etal-2024-using}. Recent methods, in particular, have leveraged the inferential and linguistic capabilities of large language models (LLMs) by integrating them with external theorem provers (TPs) to automatically verify the logical validity of explanations for NLI \citep{pan-etal-2023-logic, olausson-etal-2023-linc, quan-etal-2024-verification, dalal-etal-2024-inference}.

However, these integrated neuro-symbolic approaches still face notable challenges. First, automated theorem provers (ATP) require a machine-verifiable formal language, yet LLMs often fail to produce precise autoformalisations, underscoring their limited capacity to faithfully convert complex natural language inputs into rigorous formal representations \citep{NEURIPS2022_d0c6bc64, jiang-etal-2024-leanreasoner, quan-etal-2024-verification}. Second, syntactic errors are frequently introduced during the autoformalisation process, leading to reduced theorem-proving success rates when dealing with more complex material inferences \citep{pan-etal-2023-logic, olausson-etal-2023-linc, zhang-etal-2024-consistent}. Third, when provided with external feedback on complex explanations, LLMs often struggle to combine axioms (explanations) into cohesive proofs and effectively self-correct, limiting their effectiveness in more complex NLI settings \citep{quan-etal-2024-enhancing,quan-etal-2024-verification}.

\begin{figure*}[t]
    \centering
    \includegraphics[width=\textwidth]{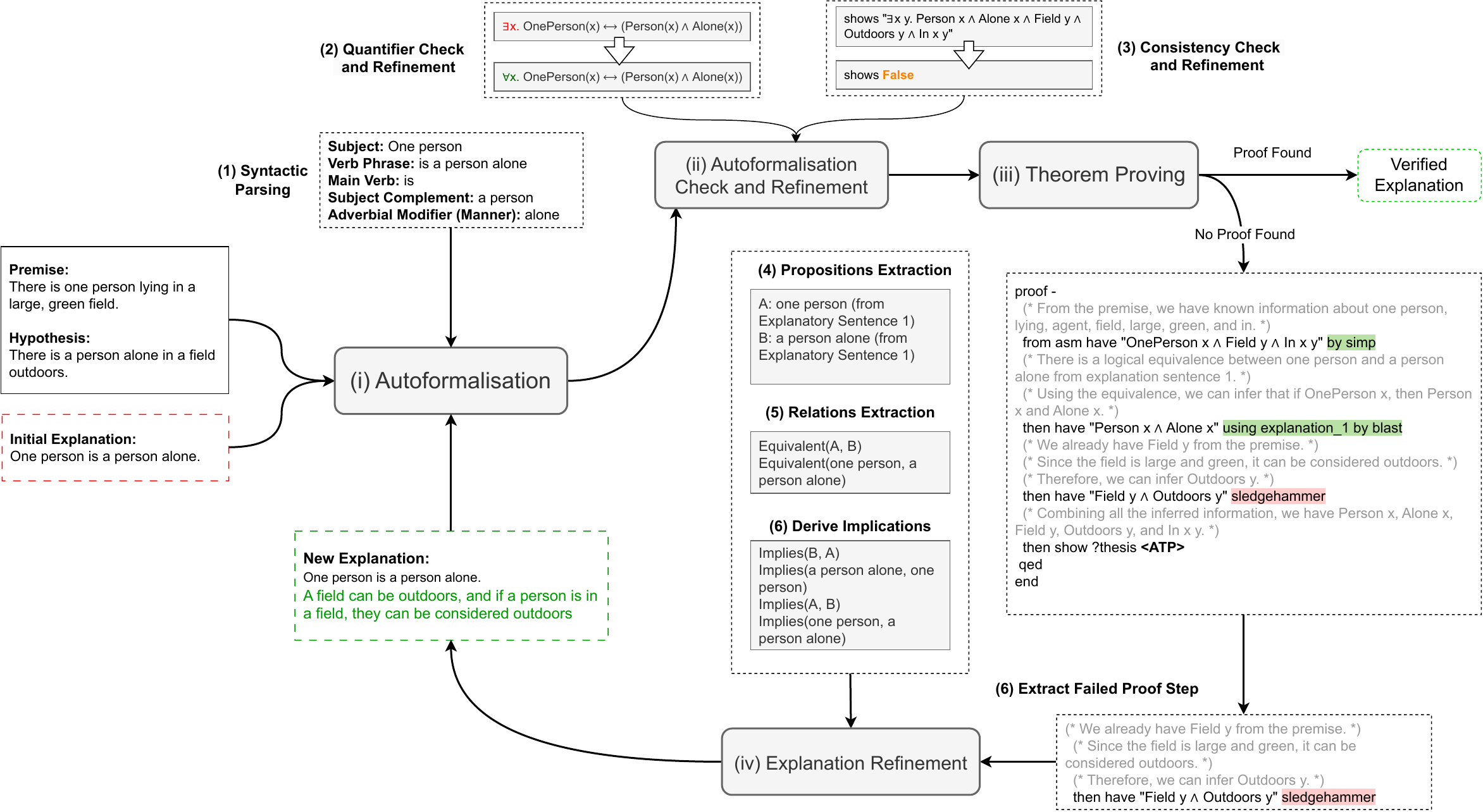}
    \caption{An illustration of our proposed interventions for improving LLM-driven theorem proving for NLI. The interventions employ different techniques including syntactic parsing, quantifier refinement, logical consistency refinement, and logical expression extraction to guide LLMs in generating more faithful and robust proof sketches for NLI and effectively refine natural language explanations. This approach provides more structured and explicit feedback by pinpointing the exact logical errors identified in the explanations.}
\label{fig:framework}
\end{figure*}

In this paper, we build upon the state-of-the-art LLM-based theorem proving framework for NLI, Explanation-Refiner \citep{quan-etal-2024-verification}. In particular, we explore methodologies to improve the faithfulness of autoformalisation and deliver a more robust way to effectively and efficiently provide logically valid explanations. We further examine how varying degrees of dataset complexity in multi-hop reasoning affect the reliability of proof step generation in LLM-Driven theorem proving. In general, we implement a neuro-symbolic framework to address the following research questions: \emph{RQ1: "To what extent can we deliver faithful autoformalisation that preserves semantic information?" RQ2: "What types of syntactic errors commonly appear in formal representations, and how effectively can state-of-the-art LLMs refine these errors?" RQ3: "Can state-of-the-art LLMs generate structured proof steps that can effectively provide feedback to refine explanations with complex sentences and logical relations?"}

To answer these questions, we investigate how to systematically leverage syntactic parsing during autoformalisation to guide LLMs generate logical representation of explanations. In addition, we define the general autoformalisation error types and use LLMs to refine these errors explicitly from the output message of a TP. Furthermore, we propose a method to extract the logical propositions, relations and implications to guide LLMs to generate proof sketches for automated theorem proving and explanation refinement. 

% Our empirical evaluation on e-SNLI \citep{NEURIPS2018_4c7a167b}, QASC \citep{Khot2019QASC}, and WorldTree \citep{jansen-etal-2018-worldtree} shows that the proposed framework identifies errors in autoformalisation and refines them with an average error reduction of 85.71\%, 83.33\%, and 81.25\%, surpassing the respective 80.65\%, 76.47\%, and 49.09\% achieved by Explanation-Refiner. Furthermore, when measuring faithfulness in autoformalisation by converting logical forms back into natural language, our approach achieves higher similarity with the original statements (i.e., 0.7944, 0.7794, and 0.5979), compared to Explanation-Refiner (i.e., 0.6737, 0.5730, and 0.3024). Additionally, the number of refined explanations produced by our framework exceeds that of Explanation-Refiner across all LLMs: raising refinement rates from 41\% to 95\%, 17\% to 90\%, and 7\% to 73\% across all three datasets. To summarise, the main contributions of this paper are:

Our empirical evaluation on e-SNLI \citep{NEURIPS2018_4c7a167b}, QASC \citep{Khot2019QASC}, and WorldTree \citep{jansen-etal-2018-worldtree} shows that the proposed framework improves the faithfulness of autoformalisation by 18.46\%, 34.2\%, 39.77\%, respectively, compared to Explanation-Refiner. Additionally, the number of refined explanations produced by our framework exceeds that of Explanation-Refiner across all LLMs: raising refinement rates from 41\% to 95\%, 17\% to 90\%, and 7\% to 73\% across all three datasets. To summarise, the main contributions of this paper are:

\begin{enumerate}
    \item We introduce Faithful-Refiner, a novel neuro-symbolic framework that provides more robust and faithful verification and refinement of explanations in NLI, surpassing existing LLM-driven theorem-proving approaches.
    \item We conduct a quantitative evaluation of explanation refinement and autoformalisation across different LLMs, achieving an average improvement of 29.5\%, 51.5\%, and 41.25\% more refined explanations, as well as 5.06\%, 6.86\%, and 32.16\% on syntactic errors reduction compared to the state-of-the-art.
    \item We adopt a range of automatic metrics to measure the quality of explanations and autoformalisation, showing that the proposed framework significantly improve the faithfulness of the autoformalisation process.
    \item We also perform a manual evaluation to assess the perceived quality of the formalised logical forms and conduct an extensive ablation study, elucidating the role of each proposed component and identifying key factors influencing automated theorem proving for NLI.
\end{enumerate}

\section{Automated Theorem Proving for Explanation-Based NLI}
In this paper, we define an \emph{explanation} $E_i$ as a set of facts $\{f_1,f_2,\ldots,f_n\}$ that establish a logically valid entailment between \emph{premises} $p_i$ and a \emph{hypothesis} $h_i$, such that $p_i \cup E_i \models h_i$ holds. 

In this work, we leverage an external theorem prover $TP$ to systematically verify these entailments in an automated manner. Specifically, given the set of input sentences $S = p_i \cup \{h_i\} \cup E_i$, we aim to build a set of logical forms $\phi = \{\Phi(s)\mid s \in S\}$, where $\Phi$ is the \emph{autoformalisation process} that converts natural language sentences into symbolic representations. From these logical forms, we construct a theory $\Theta = (A, \tau)$, where $A = \{a_1, a_2, \ldots, a_n\}$ is the set of axioms derived from formalising $E_i$, and $\tau$ is the theorem to be proven, composed of $p_i$ and $h_i$. If an automated theorem prover ($ATP$) can derive a valid proof for $\Theta$, we conclude that $E_i$ is \emph{sound and logically valid}. Otherwise, we refine $E_i$ by using the failed proof steps as feedback, iteratively generating a refined explanation $E_i'$ that ultimately leads to a valid justification.

\section{Methodology}
To effectively enhance the joint inference capabilities and robustness between LLMs and theorem provers for explanation-based NLI, we propose a novel framework to enhance three key components: autoformalisation, logical and syntactic error checking and refinement, and LLM-guided proof construction. As illustrated in Figure \ref{fig:framework}, the pipeline begins with the automated formalisation of natural language into logical representations.

Unlike the previous state-of-the-art approach (i.e., Explanation-Refiner), we begin with a syntactic parsing step that guides LLMs in translating natural language elements into a formal specification compatible with theorem provers. The LLM is prompted to automatically formalise the explanatory sentences into axioms and construct a theorem composed of assumption clauses (drawn from the premise) and a proof goal (derived from the hypothesis). After formalising the input sentences, we apply a quantifier and a logical consistency check along with a refinement process.

Similar to \citet{Jiang2022DraftSA} and \citet{quan-etal-2024-verification}, we adopt Isabelle/HOL \citep{nipkow2002isabelle} to formally verify the constructed theory. Specifically, we invoke the Sledgehammer tool \citep{Paulson2012ThreeYO} within Isabelle/HOL to call upon multiple automated theorem provers (e.g., CVC4\footnote{\url{https://cvc4.github.io/}}, Vampire\footnote{\url{https://vprover.github.io/projects.html}}), which attempt to prove the theorem derived from the translated NLI tasks. If any prover succeeds, we conclude that the explanation is logically sound, thereby confirming that the premise entails the hypothesis. 

If no proof is found, we use an LLM to extract logical propositions and relations from the natural language explanations. 
We then employ an intermediate propositional representation to derive further implications among these propositions, prompting the LLM to generate a step-by-step proof sketch—rather than having the LLM serve directly as a proof planner as in Explanation-Refiner.

Finally, we iteratively attempt to prove each sub-proof step, gathering information about failed steps, using it as feedback to prompt the LLM to generate an updated explanation to refine the logical errors identified in the previous proof sketch and start a new iteration.

\begin{figure*}
\begin{lstlisting}
(* Explanation 1: A man and woman are at the park. *)
axiomatization where
  explanation_1: "$\exists$x y z. Man x $\wedge$ Woman y $\wedge$ Park z $\wedge$ At x z $\wedge$ At y z"
theorem hypothesis:
  (* Premise: A man and woman sit on a park bench with a set of newlyweds behind *)
  assumes asm: "Man x $\wedge$ Woman y $\wedge$ ParkBench z $\wedge$ Newlyweds w $\wedge$ Sit e $\wedge$ Agent e x $\wedge$ Agent e y $\wedge$ Patient e z $\wedge$ Behind w z"
  (* Hypothesis: People outside *)
  shows "$\exists$x. People x $\wedge$ Outside x"
proof -
  (* From the premise, we have information about a man and a woman sitting on a park bench. *)
  from asm have "Man x $\wedge$ Woman y" by blast
  (* Explanation 1 states that a man and a woman are at the park. *)
  (* This implies that they are outside, as parks are typically outdoor locations. *)
  from explanation_1 have "$\exists$x y z. Man x $\wedge$ Woman y $\wedge$ Park z $\wedge$ At x z $\wedge$ At y z" by blast
  (* Since a man and a woman are at the park, they are outside. *)
  then have "People x $\wedge$ Outside x" <ATP>
  then show ?thesis <ATP>
qed
\end{lstlisting}
\caption{An example of a proof sketch constructed by the model to verify an explanation from the e-SNLI dataset. In this example, while ATPs find proofs for the first two steps using proof tactics, they fail to derive \emph{People $x \wedge$ Outside $x$} due to missing premises. The feedback provided by Isabelle is then adopted in the next iteration to refine the explanation and the proof sketch.} 
\label{proof_example}
\end{figure*}

\subsection{Isabelle/HOL Theory Generation}
Autoformalisation plays a critical role in integrating theorem provers with LLMs, especially for complex sentence structures. Similar to \citet{quan-etal-2024-verification}, we apply Neo-Davidsonian event-based semantics \citep{Parsons1990EventsIT} to formalising the natural language sentences within each aspect of an event with distinct predicates. This approach provides a robust foundation for formalising explanatory sentences while maximising content preservation \citep{maienborn2011semantics}. 

However, simply using few-shot prompting for autoformalisation does not guarantee a faithful process, which may lead to inconsistencies between the natural and formal languages expressions. To alleviate this, we begin by performing syntactic parsing via the LLMs on all provided sentences to extract their grammatical structure, identifying the main predicate-argument structure. These elements are subsequently mapped onto the agent, event action, and patient roles within a Neo-Davidsonian event semantics framework. For example, consider the sentence \emph{"The father and son kicked the ball"}. We can parse it as:
\newline
\Tree [ .S 
          [ .NP-SBJ The father and son ]
          [ .VP 
            [ .V kicked ]
            [ .NP-OBJ the ball ]
          ]
      ]
\newline
\noindent indicating that "The father and son" is the subject while "the ball" is the object. Thus we could build the Neo-Davidsonian event semantics to formalise it as:
\begin{lstlisting}[]
$\exists$xyze. (Father(x) $\wedge$ Son(y) $\wedge$ Ball(z) $\wedge$ Kicked(e) $\wedge$ Agent(e, x) $\wedge$ Agent(e, y) $\wedge$ Patient(e, z))
\end{lstlisting}
By leveraging such a process, we construct a clear representation indicating that the father and the son are the agents performing the event (kick), while the ball is the patient receiving the action, thus capturing all relevant semantic information in the transition from natural language to formal language. We then construct the Isabelle/HOL theory with axioms (explanatory sentences) and the theorem (premise and hypothesis sentences).

\subsection{Autoformalisation Critiques}
Recent studies have identified errors and inconsistencies in LLM-generated outputs as a challenge in autoformalisation and have proposed several methods \citep{pan-etal-2023-logic, zhang-etal-2024-consistent, gandarela2025inductivelearninglogicaltheories} to address them. In our work, we categorise the errors in this phase into three main dimensions: quantifier scoping error, syntax errors, and logical inconsistencies.

\paragraph{Quantifier Scoping Error}
The quantifiers indicate the scope of logical deductions. In synthetically generated datasets quantifiers are constrained to predefined settings. In contrast, in naturally occurring NL settings, incorrect quantifiers in axioms may still prove a theorem within a formal system, but when those logical forms are restated in natural language, their soundness may fail to hold in the real world. For example, one cannot declare “all animals are mammals." Thus, we introduce a \textit{quantifier check and refinement} soft-critique stage to prompt the LLM to compare the quantifiers in the logical forms against real-world knowledge, thereby avoiding any over-scoped quantifiers.

\paragraph{Syntax Errors}
Internal syntax errors, primarily those caused by missing brackets or type unification conflicts of logical variables, can often be identified through the theorem prover’s output. Once identified through a hard critique via the TP, these errors can be systematically refined or corrected by adjusting the syntax or revising type declarations. We then employ an LLM for refinement to support the systematic correction of these output errors (constrained within up to five iterations).

\paragraph{Logical Inconsistencies}
In a formal system, if contradictory or meaningless axioms are introduced, the system becomes inconsistent. By the principle of explosion (\emph{ex falso [sequitur] quodlibet}), any proposition can then be derived from such an inconsistency. To test such errors within the autoformalised axioms, we construct a modified theorem $\tau_{\text{False}}$ by replacing the conclusion of $\tau$ with "False". We then attempt to prove this modified theorem, if the TP finds a proof, it indicates a contradiction within the axioms. In this case, we use an LLM to refine the axioms and attempt to solve the contradictions.

\begin{table*}[t]
\centering
\small
\renewcommand{\arraystretch}{1.15}
\setlength{\tabcolsep}{3.5pt}
\sisetup{number-unit-product = {}}
\begin{tabular}{
@{}l
  S[table-format=3.2]   % e-SNLI Init.
  S[table-format=4.2]   % e-SNLI Final. 
  S[table-format=1.2]   % e-SNLI #Iter
  S[table-format=2.2]   % e-SNLI #Calls
  S[table-format=3.2]   % QASC Init.
  S[table-format=4.2]   % QASC Final.
  S[table-format=1.2]   % QASC #Iter
  S[table-format=2.2]   % QASC #Calls
  S[table-format=2.2]   % WorldTree Init.
  S[table-format=4.2]   % WorldTree Final.
  S[table-format=1.2]   % WorldTree #Iter
  S[table-format=2.2]   % WorldTree #Calls
@{}}
\toprule
& \multicolumn{4}{c}{\textbf{e-SNLI}}
& \multicolumn{4}{c}{\textbf{QASC}}
& \multicolumn{4}{c}{\textbf{WorldTree}} \\
\cmidrule(lr){2-5}\cmidrule(lr){6-9}\cmidrule(lr){10-13}
\textbf{}   
& \textbf{Init.} & \textbf{Final} & \textbf{\#Iter} & \textbf{\#Calls}
& \textbf{Init.} & \textbf{Final} & \textbf{\#Iter} & \textbf{\#Calls}
& \textbf{Init.} & \textbf{Final} & \textbf{\#Iter} & \textbf{\#Calls} \\
\midrule
\multicolumn{13}{@{}l}{\textit{\textbf{Explanation-Refiner}}} \\
\midrule
Llama3.1-70b 
 & \SI{23}{\percent} & \SI{51}{\percent} & 4.08 & 34.56
 & \SI{4}{\percent} & \SI{18}{\percent} & 4.07 & 37.49
 & \SI{2}{\percent} & \SI{15}{\percent} & 5.23 & 51.61 \\
GPT-4o-mini 
 & \SI{13}{\percent} & \SI{30}{\percent} & 3.65 & 32.55
 & \SI{3}{\percent} & \SI{20}{\percent} & 5.12 & 44.84
 & \SI{0}{\percent} & \SI{4}{\percent} & 5.00 & 46.12 \\
GPT-4o
 & \SI{31}{\percent} & \textbf{\underline{\SI{71}{\percent}}} & 3.62 & 32.34
 & \SI{4}{\percent} & \SI{26}{\percent} & 4.35 & 38.45
 & \SI{2}{\percent} & \SI{13}{\percent} & \textbf{\underline{4.18}} & 39.26 \\
Deepseek-V3 
 & \SI{25}{\percent} & \SI{69}{\percent} & \textbf{\underline{2.82}} & 27.74
 & \SI{4}{\percent} & \textbf{\underline{\SI{38}{\percent}}} & \textbf{\underline{3.71}} & 35.97
 & \SI{3}{\percent} & \textbf{\underline{\SI{31}{\percent}}} & 4.52 & 42.64 \\
\midrule
\multicolumn{13}{@{}l}{\textit{\textbf{Our Approach}}} \\
\midrule
Llama3.1-70b 
 & \SI{36}{\percent} & \SI{78}{\percent} & 2.38 & 16.28
 & \SI{11}{\percent} & \SI{68}{\percent} & 2.90 & 25.40
 & \SI{6}{\percent} & \SI{52}{\percent} & 4.62 & 35.72 \\
GPT-4o-mini
 & \SI{32}{\percent} & \SI{77}{\percent} & 2.27 & 16.62
 & \SI{12}{\percent} & \SI{71}{\percent} & 3.35 & 27.10
 & \SI{5}{\percent} & \SI{47}{\percent} & 4.75 & 36.50 \\
GPT-4o
 & \SI{39}{\percent} & \SI{89}{\percent} & 1.54 & 10.24
 & \SI{10}{\percent} & \SI{79}{\percent} & 3.22 & 22.32
 & \SI{9}{\percent} & \SI{56}{\percent} & 3.86 & 26.16 \\
Deepseek-V3 
& \SI{41}{\percent} & \textbf{\underline{\SI{95}{\percent}}} & \textbf{\underline{1.50}} & \textbf{\underline{9.52}}
& \SI{17}{\percent} & \textbf{\underline{\SI{90}{\percent}}} & \textbf{\underline{2.53}} & \textbf{\underline{20.18}}
& \SI{7}{\percent} & \textbf{\underline{\SI{73}{\percent}}} & \textbf{\underline{3.55}} & \textbf{\underline{25.30}}
 \\
\midrule
\end{tabular}
\caption{Comparison of our approach with Explanation-Refiner on different LLMs across three datasets. Init. represents the number of explanations that are initially verified as logically valid. Final indicates the number of explanations that are refined within a maximum of 10 iterations, while \#Iter indicates the average iteration required to refine an explanation. \#Calls shows the average number of LLM calls needed to fully refine an explanation.}
\label{tab:quantitative-explanation}
\end{table*}

\subsection{Proof, Verification and Refinement}
After autoformalisation checking and refinement, we employ the theorem prover $TP$ to verify the logical validity of the axioms and determine whether $A \models \tau$ holds. We first use the Sledgehammer tool in Isabelle/HOL for ATPs to automatically find a proof of the theorem. If a proof is found, we extract all possible proofs from Sledgehammer's results and state that the explanation is logically valid. If Sledgehammer fails to find a proof, we construct a proof sketch to attempt a step-by-step proving using ATPs based on a set of logical interpretations. 

\paragraph{Logical Propositions, Relations and Implications}
\citet{liu2025logicofthoughtinjectinglogiccontexts} employ logical expressions to guide LLMs and mitigate information loss in intermediate reasoning processes. Similarly, we begin with a logical proposition extraction step. In this step, we use an LLM to extract logical propositions and relations from the explanation $E_i$. Consider the following extracted logical relations as an example: A: it is raining; B:  the grass is wet; C: kids can play outside; D: kids are happy as well as the following logical relations: $A \rightarrow B$ (if it is raining, the grass is wet) and $B \rightarrow \neg C$ (if the grass is wet, kids cannot play outside). Next, we leverage the extracted logical relations using a SymPy-based propositional-level representation \citep{Meurer:2017yhf} \footnote{\href{Sympy:}{https://www.sympy.org/en/index.html}} to derive additional implications based on formal logical laws. For instance, from the example above, SymPy can deduce $A \rightarrow \neg C$ (if it is raining, kids cannot play outside). Algorithm \ref{algorithm_1} shows the implementation of SymPy to find derived logical implications.

\paragraph{Proof Sketch}
By combining the logical propositions, relations, and these derived implications, the LLM can construct a step-by-step guided proof sketch that establishes a logical reasoning chain to prove the goal. As shown in Figure \ref{proof_example}, the comments partially indicates how the logical expression guides LLMs to build the step-wise proof steps, while we replace the proof tactics with <ATP>, which uses Sledgehammer (Isabelle’s automated theorem proving tool) to search for proofs. In Isabelle/HOL, proof tactics are commands that systematically decompose complex proofs into simpler sub-goals, automating routine steps such as simplification. Typically, one is asked to prove a statement X given assumptions Y by using proof tactics Z, where Z includes commands like simp (for simplification), auto (for automatic reasoning), and blast (for first-order reasoning). These tactics instruct Isabelle’s proof engine on how to process a proof step by applying appropriate rules, simplifications, or other reasoning methods. Once Sledgehammer finds a proof, only the explanatory sentences (axioms) used in that proof are retained as the final refined explanation. For example, if the proof is written as ``assms explanation\_1 explanation\_2 by blast'', then only explanation 1 and explanation 2 constitute the minimal set of explanatory sentences required to entail the hypothesis.

\paragraph{Explanation Refinement}
If the automated theorem prover fails or finds no proofs in a previous proof step, we extract that proof step along with the proof strategy from the comments part as feedback to prompt the LLM to refine the logical error (i.e., missing premises) of the related explanatory sentences and process into next iteration to iteratively verify and refine the explanation. After the explanation refinement, we drop any explanatory sentences that are not included in the proof, as they are deemed unnecessary to deriving the hypothesis from the premise and then proceed to the next iteration cycle. We followed the same prompts used in Explanation-Refiner \citep{quan-etal-2024-verification} for autoformalisation. Prompts used for syntactic parsing, quantifier refinement, logical consistency and proof steps generation are reported in Appendix \ref{appendix:prompt}.

\section{Empirical Evaluation}

\subsection{Datasets and Models}
We conducted experiments with four state-of-the-art LLMs within the proposed framework: GPT-4o \citep{DBLP:journals/corr/abs-2303-08774}, GPT-4o-mini \citep{DBLP:journals/corr/abs-2303-08774}, Llama3.1-70b \citep{grattafiori2024llama3herdmodels}, Deepseek-V3 \citep{deepseekai2024deepseekv3technicalreport}. Following \citet{quan-etal-2024-verification}, we applied three sampled NLI datasets of e-SNLI \citep{NEURIPS2018_4c7a167b}, QASC \citep{Khot2019QASC}, and WorldTree \citep{jansen-etal-2018-worldtree}  each comprising 100 instances. We compare our approach with Explanation-Refiner \cite{quan-etal-2024-verification}, a state-of-the-art LLM-driven theorem prover for NLI that adopts a similar pipeline but without incorporating the specific strategies for guiding autoformalisation via syntactic parsing, performing consistency and quantification checks, and guide refinement via proof sketches and explicit implication derivation. 

\begin{figure*}[htbp]
\centering
\begin{subfigure}{.32\textwidth} 
  \centering
  \includegraphics[width=\linewidth]{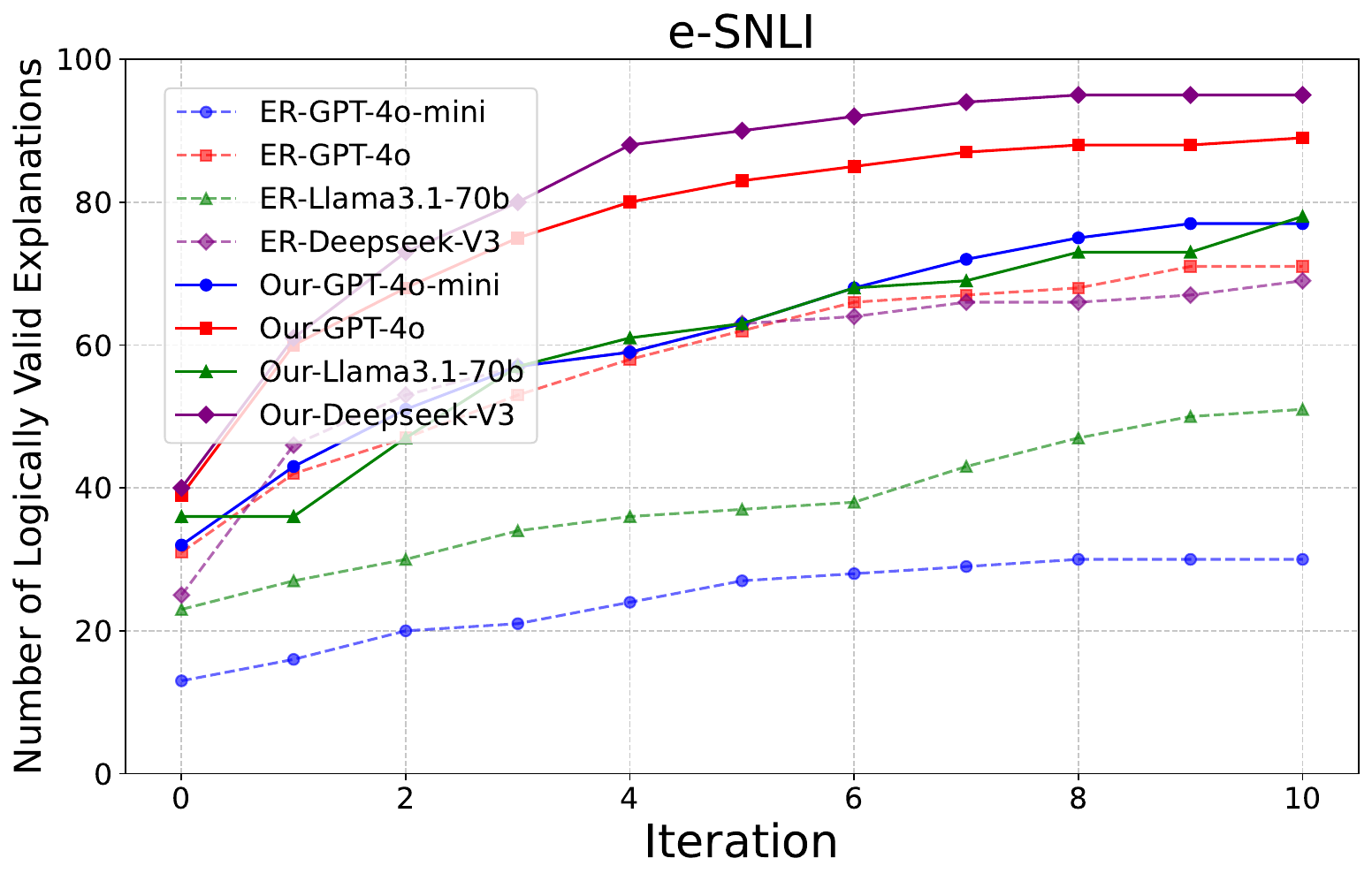}
  \caption{}
  \label{fig:sub2_overall_esnli}
\end{subfigure}
\hfill
\begin{subfigure}{.32\textwidth} 
  \centering
  \includegraphics[width=\linewidth]{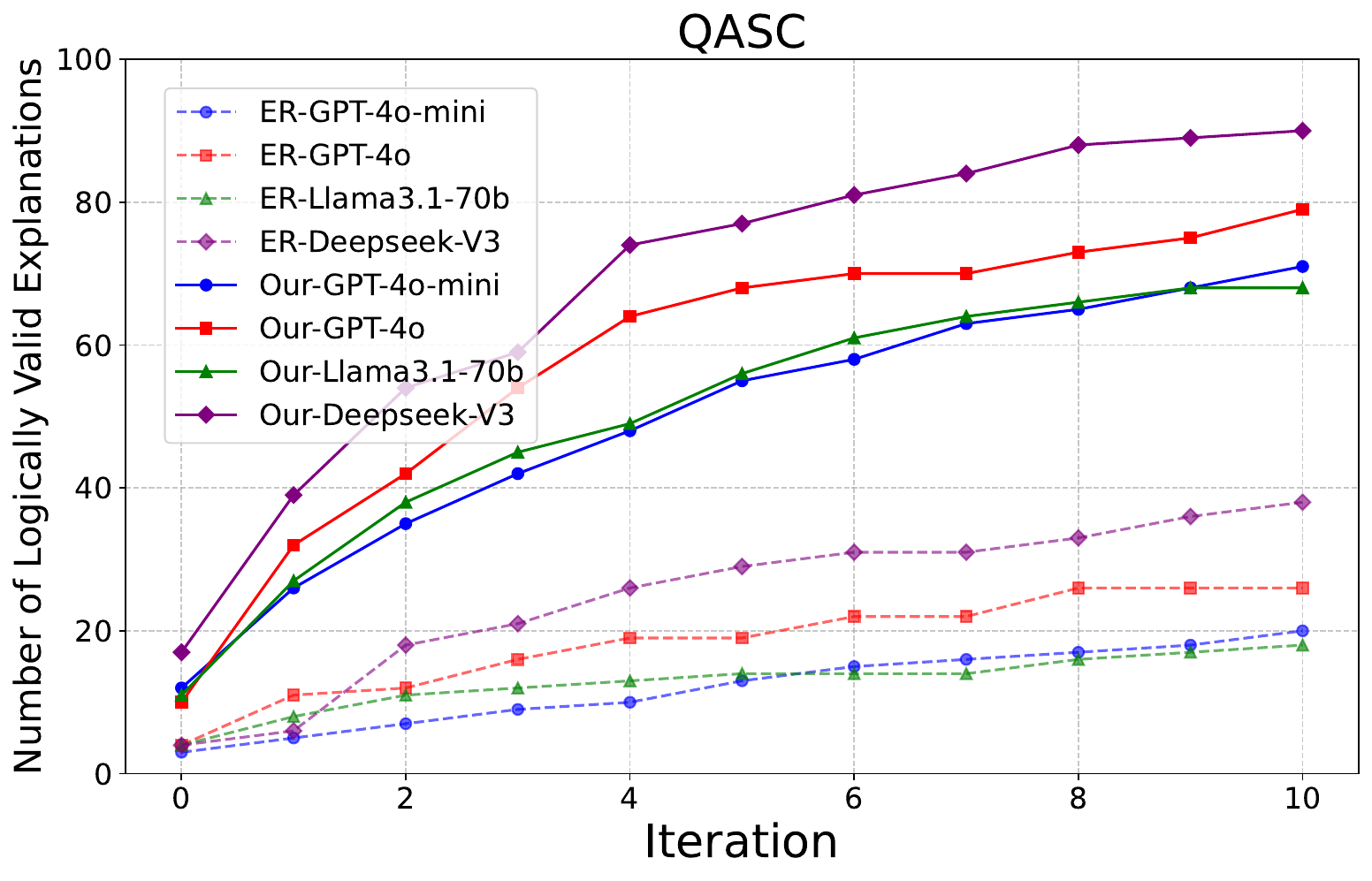}
  \caption{}
  \label{fig:sub2_overall_qasc}
\end{subfigure}
\hfill
\begin{subfigure}{.32\textwidth} 
  \centering
  \includegraphics[width=\linewidth]{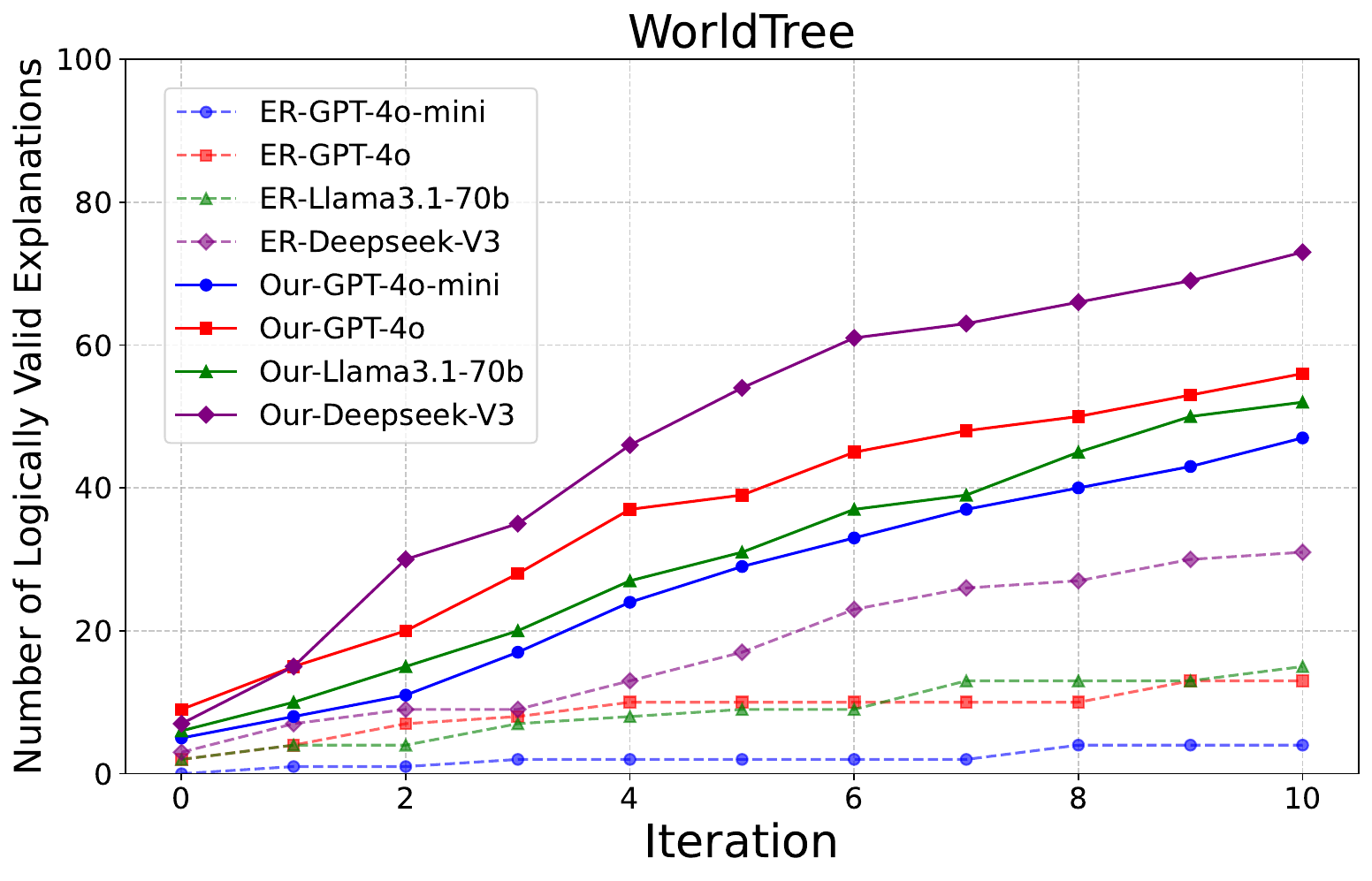}
  \caption{}
  \label{fig:sub2_overall_worldtree}
\end{subfigure}
\begin{subfigure}{.32\textwidth} 
  \centering
  \includegraphics[width=\linewidth]{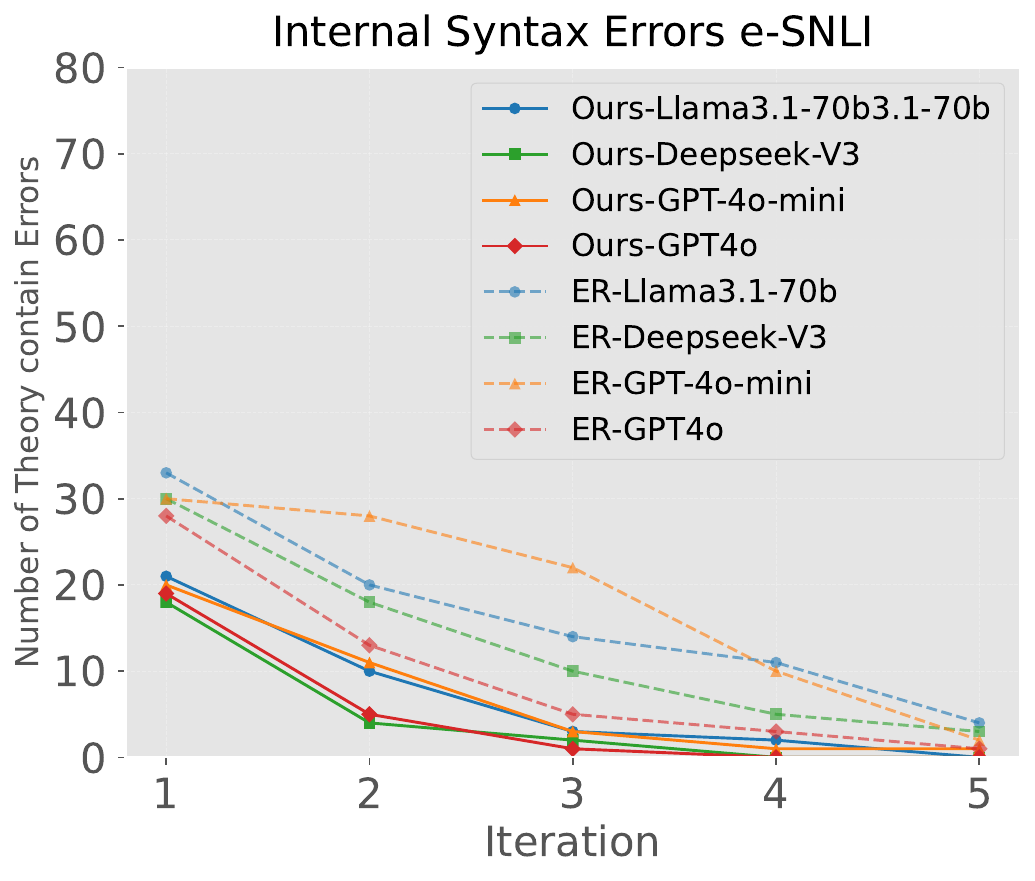}
  \caption{}
 \label{fig:sub2_syntax_esnli}
\end{subfigure}
\hfill
\begin{subfigure}{.32\textwidth} 
  \centering
  \includegraphics[width=\linewidth]{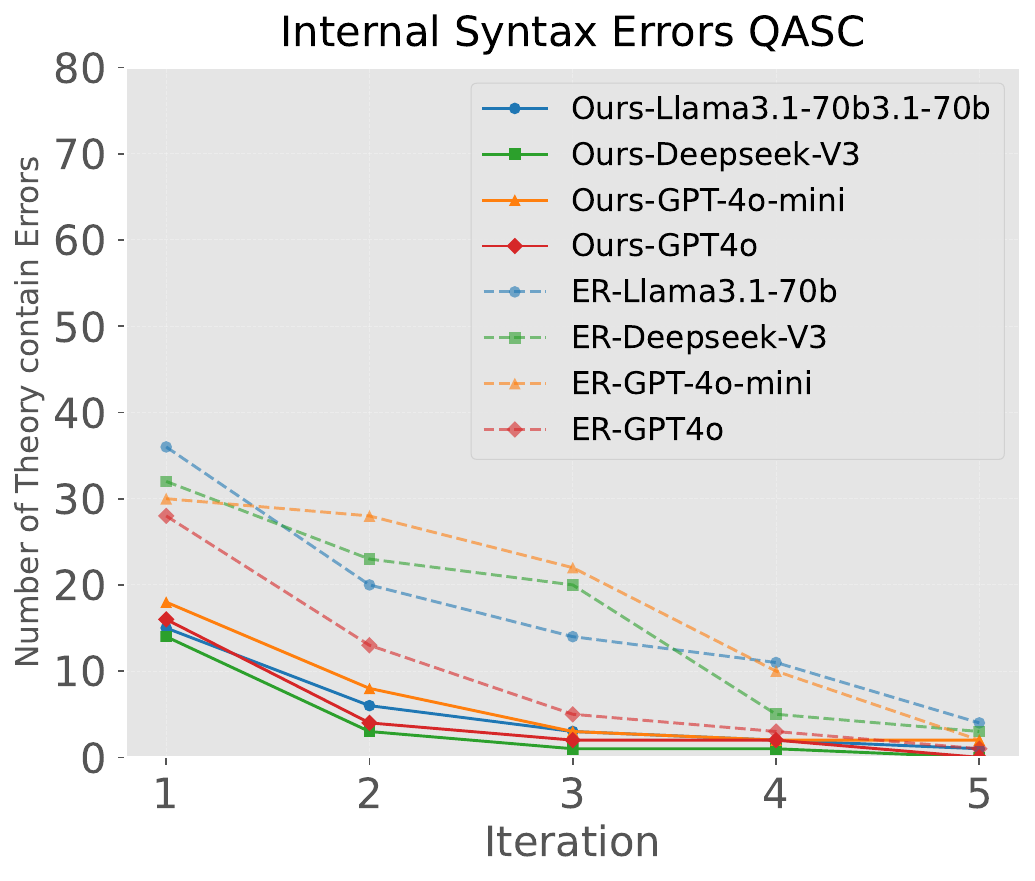}
  \caption{}
  \label{fig:sub2_syntax_qasc}
\end{subfigure}
\hfill
\begin{subfigure}{.32\textwidth} 
  \centering
  \includegraphics[width=\linewidth]{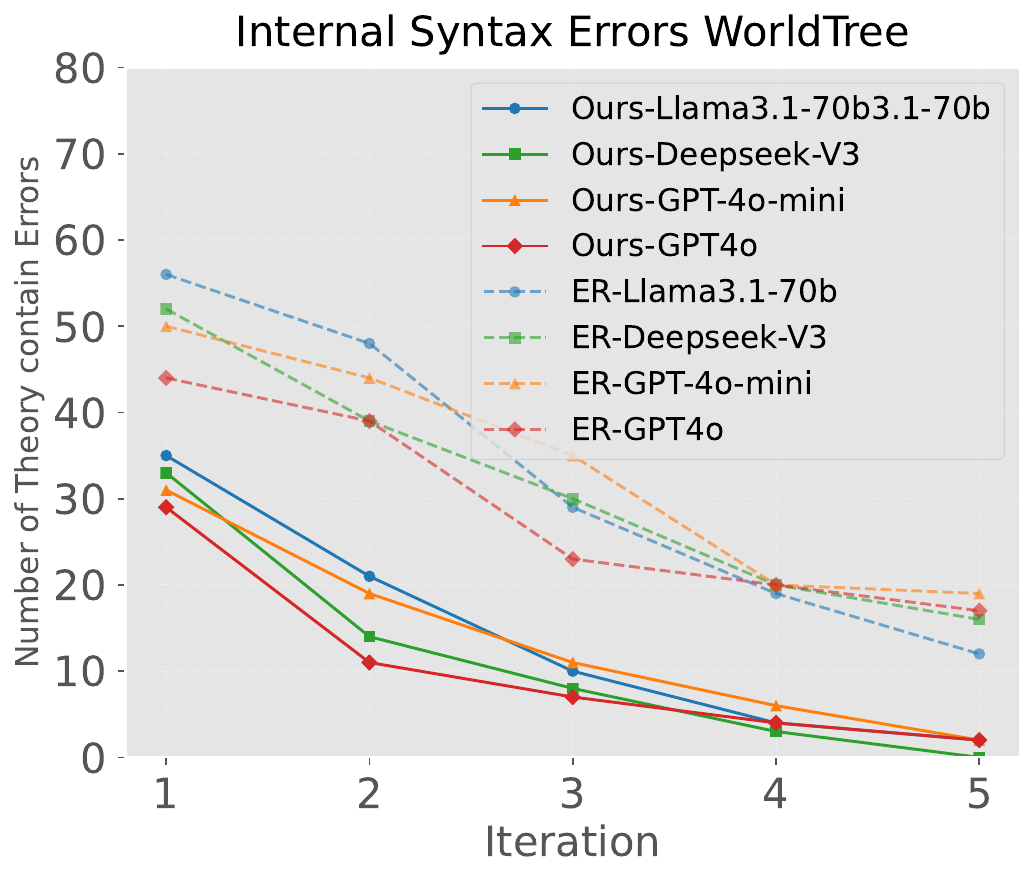}
  \caption{}
  \label{fig:sub2_syntax_worldtree}
\end{subfigure}
\caption{Top -- Number of logically valid explanations at each refinement iteration. Bottom -- Number of theories that contain internal syntactic errors at each syntax error refinement stage.}
\label{fig:overall_results_explanation_syntax_error}
\end{figure*}

\begin{table*}[t]
\centering
\tiny
\renewcommand{\arraystretch}{1.55}
\setlength{\tabcolsep}{5pt}
\sisetup{number-unit-product = {}}
\begin{tabular}{
@{}l
  S[table-format=3.2]   % e-SNLI Init.
  S[table-format=4.2]   % e-SNLI Final. 
  S[table-format=1.2]   % e-SNLI #Iter
  S[table-format=3.2]   % QASC Init.
  S[table-format=4.2]   % QASC Final.
  S[table-format=1.2]   % QASC #Iter
  S[table-format=2.2]   % WorldTree Init.
  S[table-format=4.2]   % WorldTree Final.
  S[table-format=1.2]   % WorldTree #Iter
@{}}
\toprule
& \multicolumn{3}{c}{\textbf{e-SNLI}}
& \multicolumn{3}{c}{\textbf{QASC}}
& \multicolumn{3}{c}{\textbf{WorldTree}} \\
\cmidrule(lr){2-4}\cmidrule(lr){5-7}\cmidrule(lr){8-10}
\textbf{}   
& \textbf{Init.} & \textbf{Final} & \textbf{\#Iter}
& \textbf{Init.} & \textbf{Final} & \textbf{\#Iter}
& \textbf{Init.} & \textbf{Final} & \textbf{\#Iter} \\
\midrule
\multicolumn{10}{@{}l}{\textit{\textbf{Ablations on our approach}}} \\
\midrule
GPT-4o (- \textbf{logical relations})
 & \SI{34}{\percent} & \SI{74}{\percent(-\textbf{\underline{15}}\percent)} & 2.24
 & \SI{12}{\percent} & \SI{58}{\percent(-21\percent)} & 3.46
 & \SI{6}{\percent} & \SI{38}{\percent(-18\percent)} & 4.36 \\
GPT-4o (- \textbf{detailed feedback})
 & \SI{35}{\percent} & \SI{83}{\percent(-6\percent)} & 2.86
 & \SI{13}{\percent} & \SI{56}{\percent(-\textbf{\underline{23}}\percent)} & 4.45
 & \SI{5}{\percent} & \SI{17}{\percent(-\textbf{\underline{39}}\percent)} & 6.46 \\
GPT-4o  (- refine quantifiers)
 & \SI{34}{\percent} & \SI{87}{\percent(-2\percent)} & 1.63
 & \SI{14}{\percent} & \SI{83}{\percent(+4\percent)} & 2.89
 & \SI{7}{\percent} & \SI{49}{\percent(-7\percent)} & 3.65 \\
GPT-4o  (- \textbf{refine syntax errors})
 & \SI{21}{\percent} & \SI{74}{\percent(-\textbf{\underline{15}}\percent)} & 2.34
 & \SI{5}{\percent} & \SI{58}{\percent(-21\percent)} & 4.11
 & \SI{2}{\percent} & \SI{24}{\percent(-32\percent)} &6.48 \\
\hdashline[3pt/4pt]
Deepseek-V3  (- logical relations)
 & \SI{39}{\percent}  & \SI{89}{\percent(-6\percent)} & 1.68
 & \SI{16}{\percent} & \SI{77}{\percent(-13\percent)} & 2.64
 & \SI{10}{\percent} & \SI{58}{\percent(-15\percent)} & 4.01 \\
Deepseek-V3  (- \textbf{detailed feedback})
 & \SI{31}{\percent} & \SI{86}{\percent(-9\percent)} & 3.22
 & \SI{22}{\percent} & \SI{69}{\percent(-21\percent)} & 4.12
 & \SI{6}{\percent} & \SI{41}{\percent(-\textbf{\underline{32}}\percent)} & 6.13 \\
Deepseek-V3  (- refine quantifiers)
 & \SI{34}{\percent} & \SI{96}{\percent(+1\percent)} & 1.64
 & \SI{14}{\percent} & \SI{93}{\percent(+3\percent)} & 1.89
 & \SI{6}{\percent} & \SI{70}{\percent(-3\percent)} & 3.23 \\
Deepseek-V3  (- \textbf{refine syntax errors})
 & \SI{28}{\percent} & \SI{77}{\percent(-\textbf{\underline{18}}\percent)} & 2.69
 & \SI{12}{\percent} & \SI{68}{\percent(-\textbf{\underline{22}}\percent)} & 2.84
 & \SI{4}{\percent} & \SI{46}{\percent(-27\percent)} &5.32 \\
 \midrule
\end{tabular}

\begin{tabular}{
@{}l
  S[table-format=3.2]   
  S[table-format=4.2]   
  S[table-format=2.2]   
  S[table-format=3.2]   
  S[table-format=4.2]   
  S[table-format=2.2]   
  S[table-format=2.2]   
  S[table-format=4.2]   
  S[table-format=2.2]   
@{}}
\toprule
& \multicolumn{3}{c}{\textbf{e-SNLI}}
& \multicolumn{3}{c}{\textbf{QASC}}
& \multicolumn{3}{c}{\textbf{WorldTree}} \\
\cmidrule(lr){2-4}\cmidrule(lr){5-7}\cmidrule(lr){8-10}
\textbf{}   
& \textbf{V.} & \textbf{I.} & \textbf{Q.}
& \textbf{V.} & \textbf{I.} & \textbf{Q.}
& \textbf{V.} & \textbf{I.} & \textbf{Q.} \\
\midrule
\multicolumn{10}{@{}l}{\textit{\textbf{Explanation-Refiner}}} \\
\midrule
GPT-4o
  & \textbf{\underline{\SI{9}{\percent}}} & \textbf{\underline{\SI{3}{\percent}}} & \textbf{\underline{\SI{6}{\percent}}}
 & \textbf{\underline{\SI{18}{\percent}}} & \textbf{\underline{\SI{9}{\percent}}} & \textbf{\underline{\SI{18}{\percent}}}
 & \textbf{\underline{\SI{33}{\percent}}} & \textbf{\underline{\SI{8}{\percent}}} & \textbf{\underline{\SI{16}{\percent}}} \\
Deepseek-V3 
  & \SI{27}{\percent} & \textbf{\underline{\SI{3}{\percent}}} & \SI{10}{\percent}
 & \SI{34}{\percent} & \SI{9}{\percent} & \SI{25}{\percent}
 & \SI{44}{\percent} & \SI{23}{\percent} & \SI{31}{\percent} \\
\midrule
\multicolumn{10}{@{}l}{\textit{\textbf{Ablations}}} \\
\midrule
GPT-4o  (- refine quantifiers)
  & \SI{9}{\percent} & \SI{2}{\percent} & \SI{10}{\percent(+4\percent)}
 & \SI{13}{\percent} & \SI{2}{\percent} & \SI{23}{\percent(+5\percent)}
 & \SI{38}{\percent} & \SI{6}{\percent} & \SI{19}{\percent(+6\percent)} \\
GPT-4o  (- refine syntax errors)
 & \SI{16}{\percent(+7\percent)} & \SI{1}{\percent} & \SI{3}{\percent}
 & \SI{38}{\percent(+\textbf{\underline{20}}\percent)} & \SI{3}{\percent} & \SI{7}{\percent}
 & \SI{56}{\percent(+\textbf{\underline{23}}\percent)}& \SI{3}{\percent} & \SI{16}{\percent} \\
\hdashline[3pt/4pt]  
Deepseek-V3  (- refine quantifiers)
 & \SI{25}{\percent} & \SI{5}{\percent} & \SI{16}{\percent(+\textbf{\underline{6}}\percent)}
 & \SI{31}{\percent} & \SI{14}{\percent} & \SI{35}{\percent(+\textbf{\underline{10}}\percent)}
 & \SI{32}{\percent} & \SI{11}{\percent} & \SI{38}{\percent(+\textbf{\underline{7}}\percent)}\\
Deepseek-V3  (- refine syntax errors)
 & \SI{38}{\percent(+\textbf{\underline{11}}\percent)} & \SI{4}{\percent} & \SI{11}{\percent}
 & \SI{51}{\percent(+17\percent)} & \SI{9}{\percent} & \SI{15}{\percent}
 & \SI{67}{\percent(+\textbf{\underline{23}}\percent)} & \SI{21}{\percent} & \SI{27}{\percent} \\
\bottomrule
\end{tabular}
\caption{Top -- Ablation study on the impacts of removing components from the overall architecture. Bottom -- Comparison of manually evaluated variable, implication, and quantifier errors in the autoformalisation process from a randomly sampled set of 100 Isabelle/HOL theories across all iterations for each LLM.}
\label{tab:ablation_4o_deepseek}
\end{table*}

\begin{figure*}[h!]
\centering
\begin{subfigure}{.32\textwidth} 
  \centering
  \includegraphics[width=\linewidth]{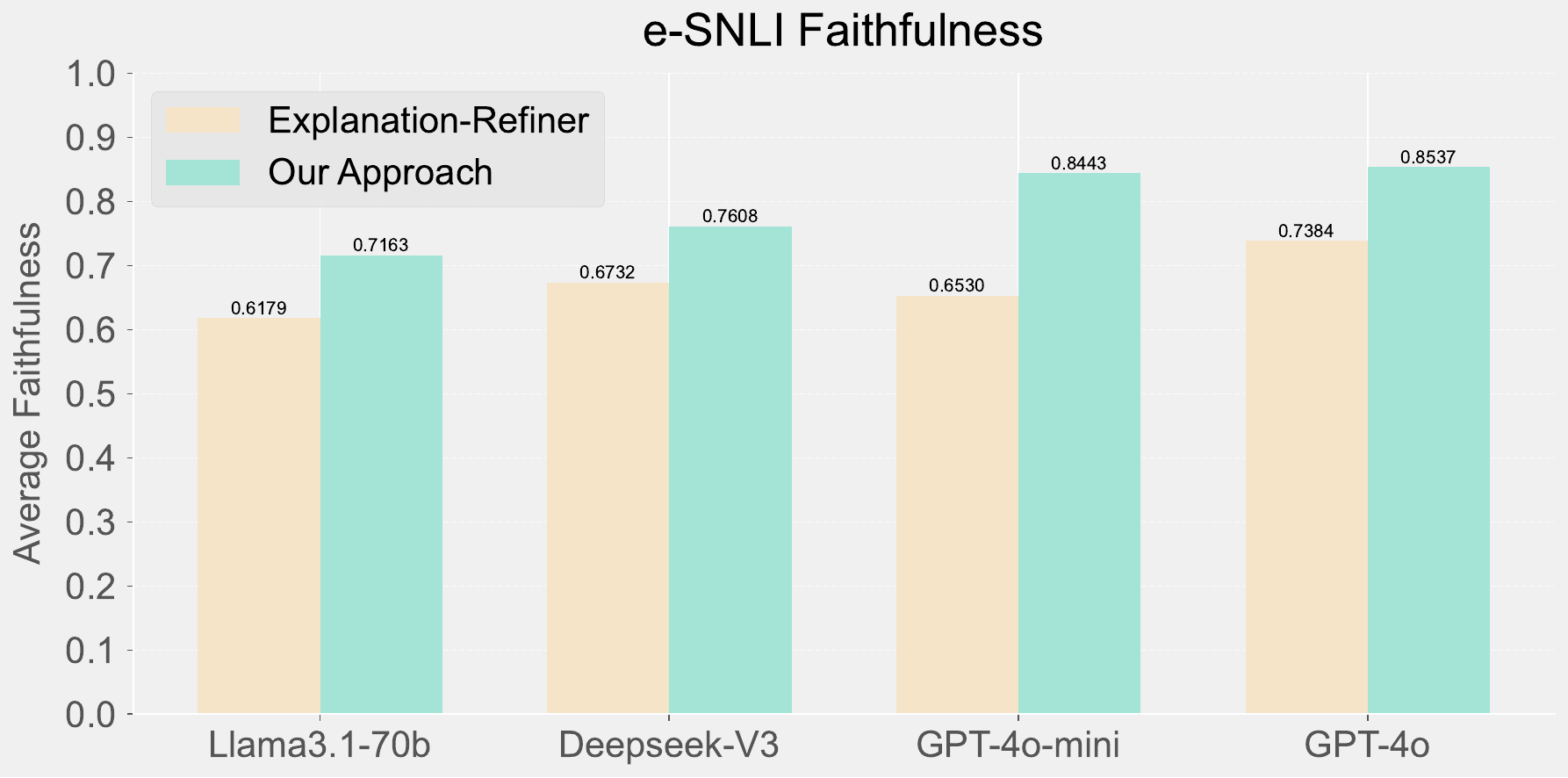}
  \caption{}
  \label{fig:sub1_autoformalisation_faithfulness}
\end{subfigure}
\hfill
\begin{subfigure}{.32\textwidth} 
  \centering
  \includegraphics[width=\linewidth]{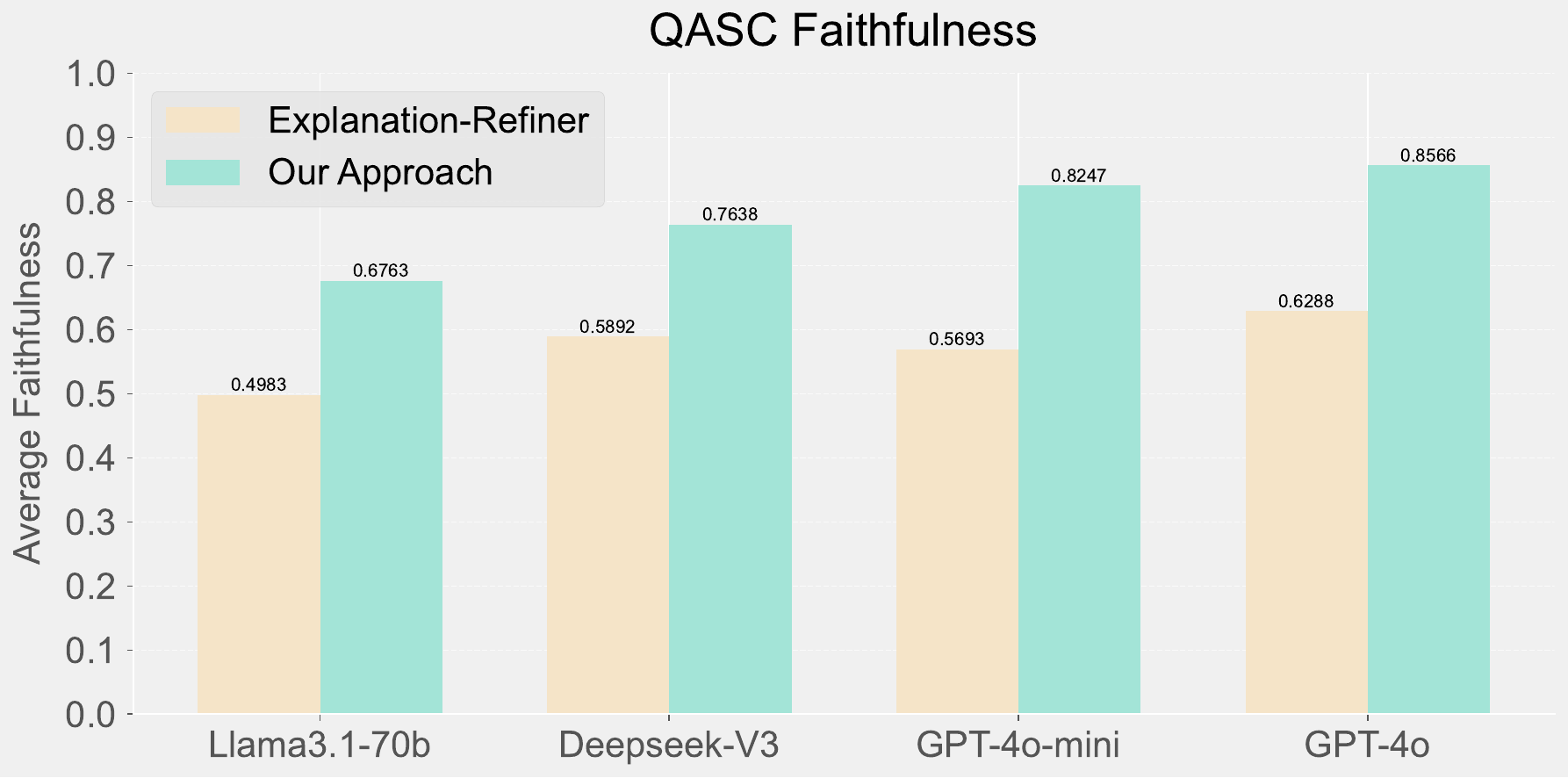}
  \caption{}
  \label{fig:sub2_autoformalisation_faithfulness}
\end{subfigure}
\hfill
\begin{subfigure}{.32\textwidth} 
  \centering
  \includegraphics[width=\linewidth]{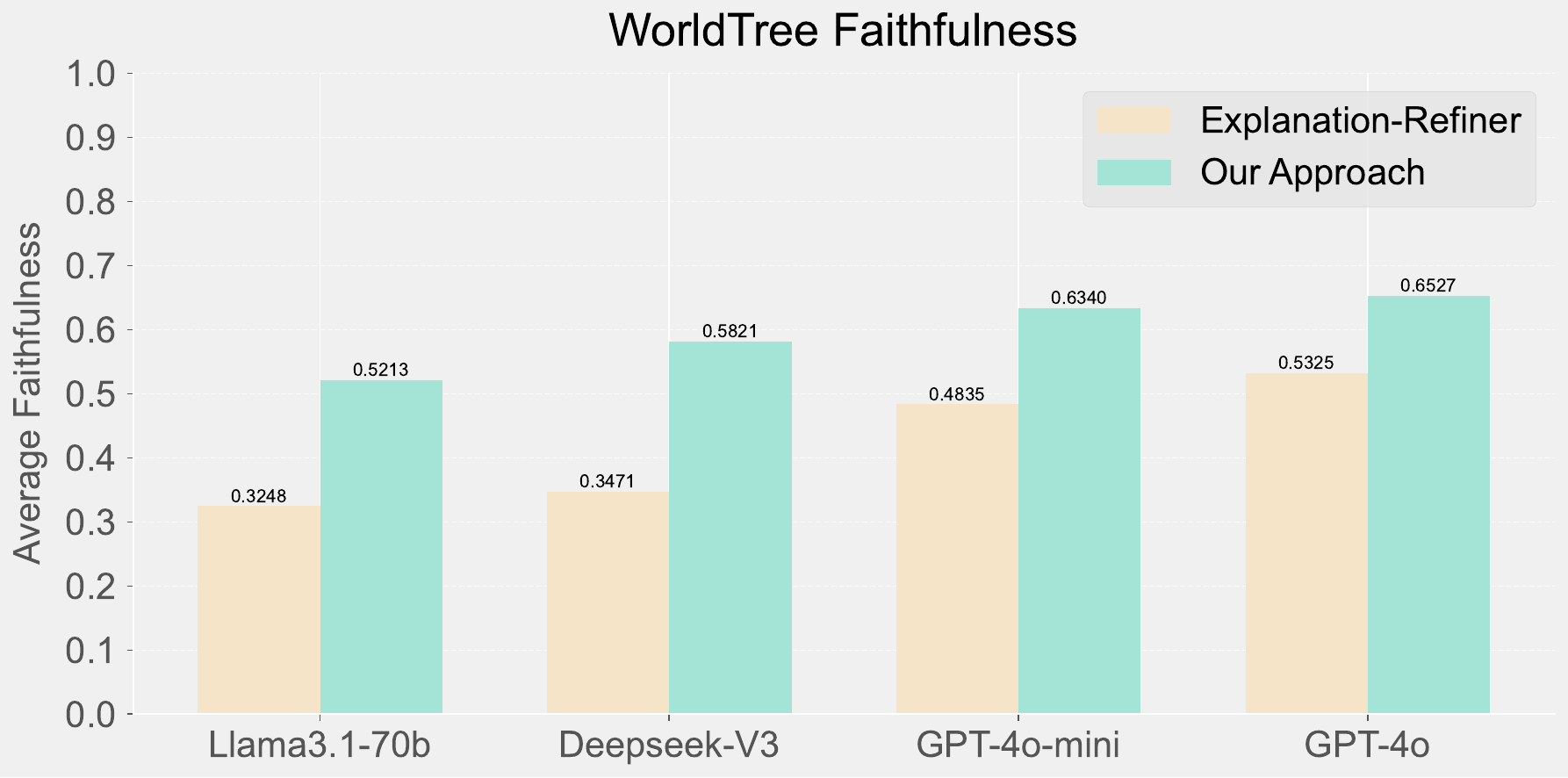}
  \caption{}
\end{subfigure}
\begin{subfigure}{.32\textwidth} 
  \centering
  \includegraphics[width=\linewidth]{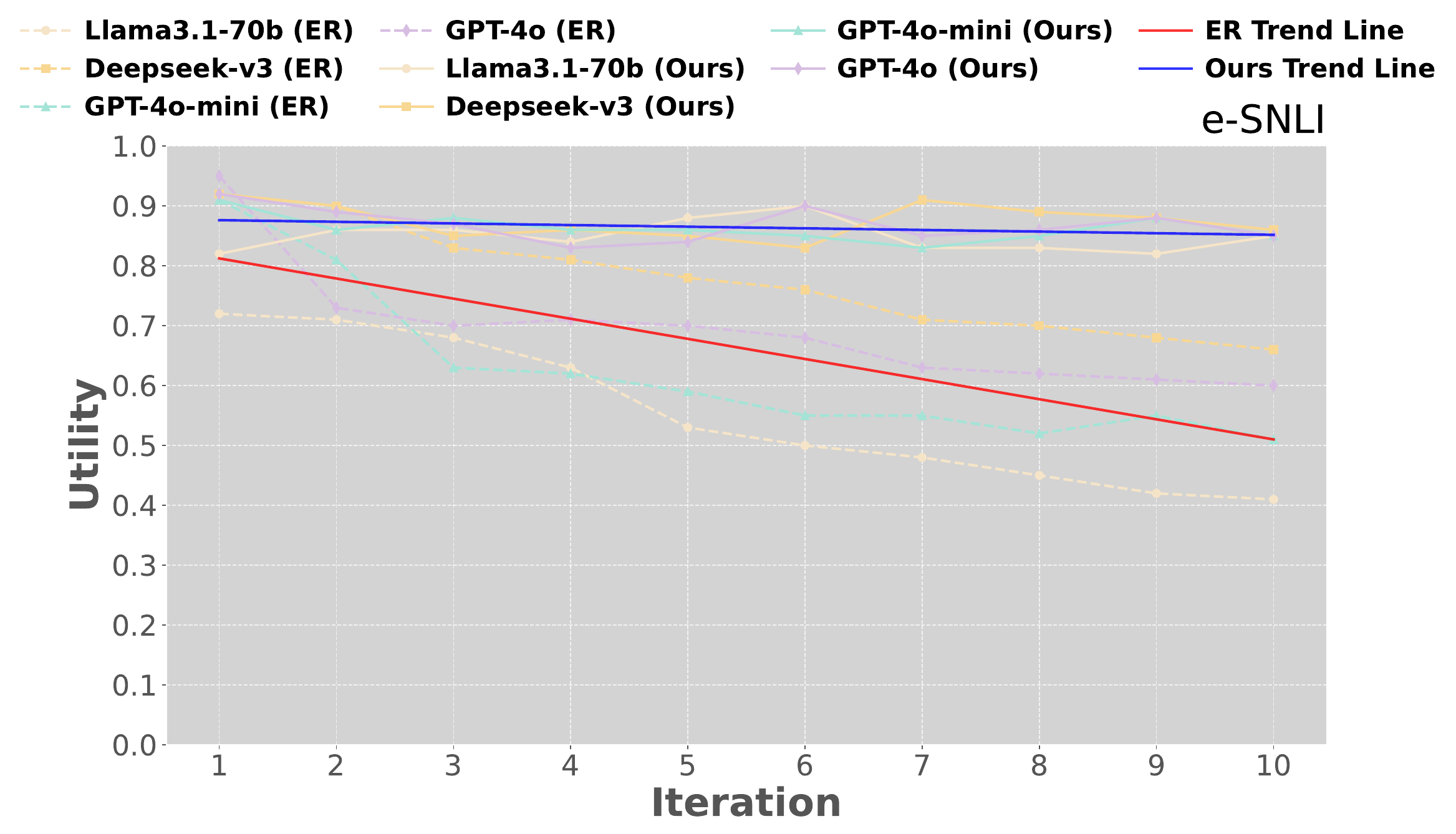}
  \caption{}
  \label{fig:sub1_utility}
\end{subfigure}
\hfill
\begin{subfigure}{.32\textwidth} 
  \centering
  \includegraphics[width=\linewidth]{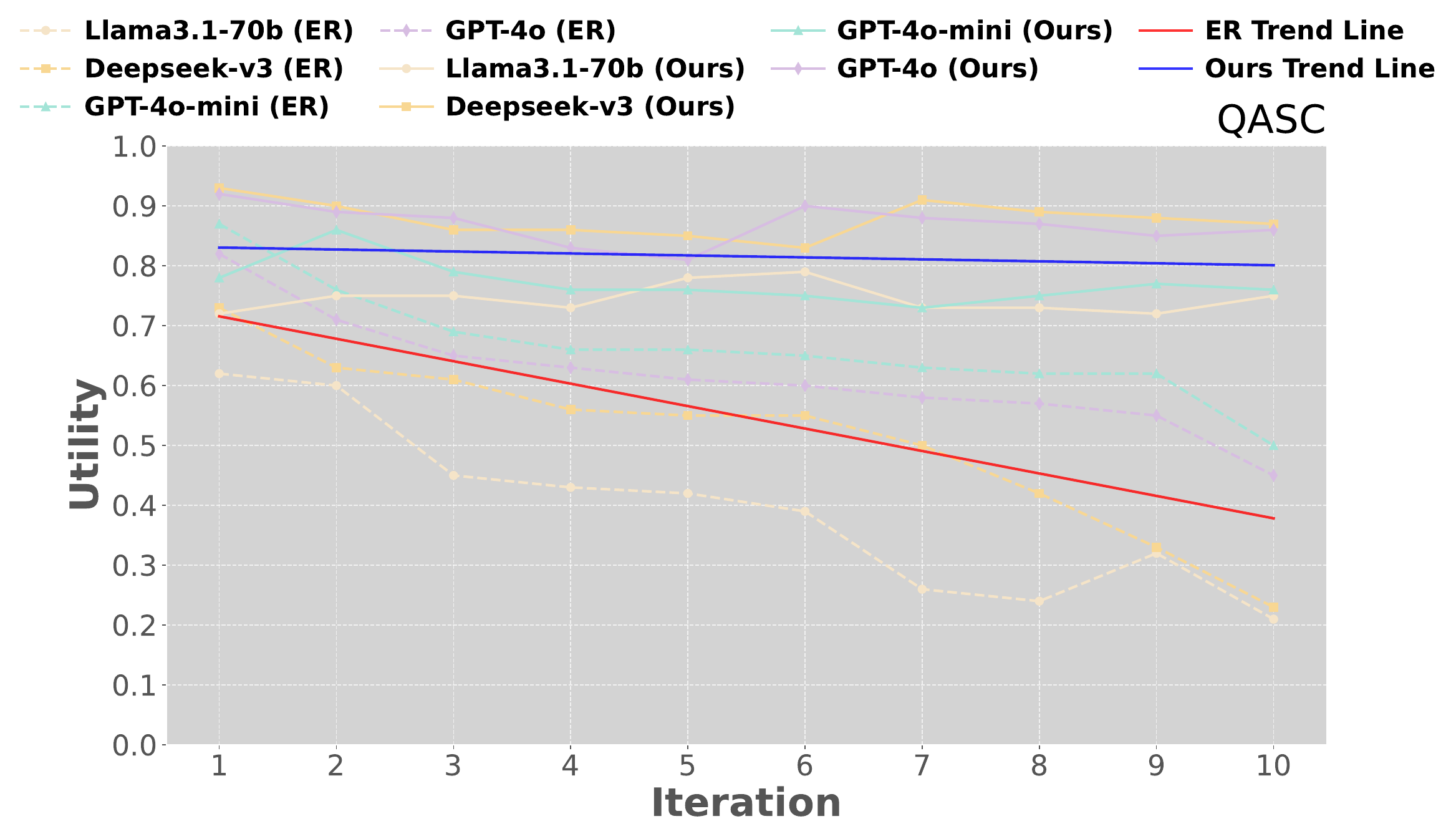}
  \caption{}
  \label{fig:sub2_utility}
\end{subfigure}
\hfill
\begin{subfigure}{.32\textwidth} 
  \centering
  \includegraphics[width=\linewidth]{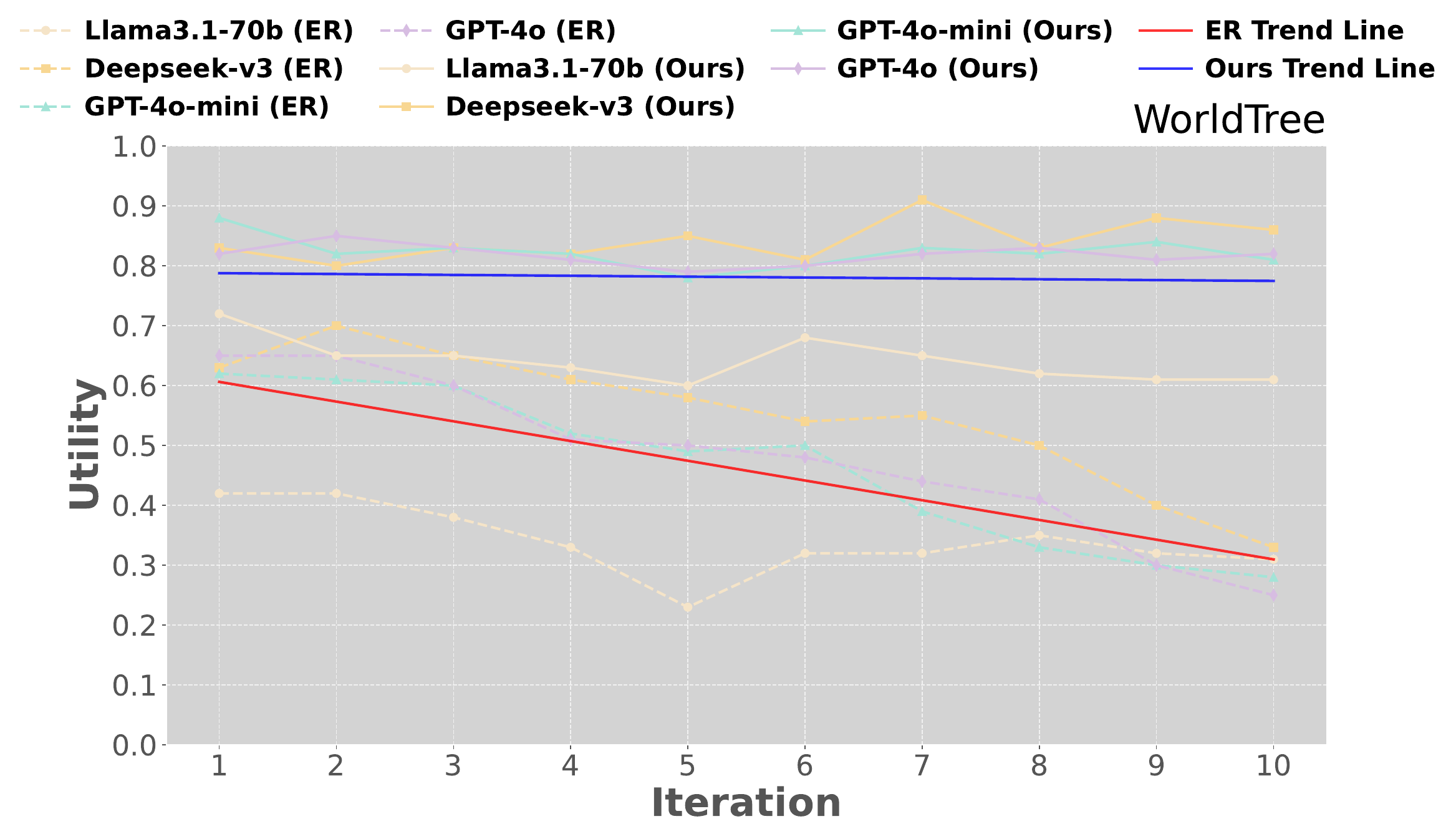}
  \caption{}
\end{subfigure}
\caption{Top -- The average faithfulness of the autoformalisation process across different LLMs. Bottom -- The utility of explanation at different refinement iterations. A higher utility indicating the newly refined explanation are more likely be used in the proof of next iteration.}
\label{fig:faithfulness_utility}
\end{figure*}

\subsection{Results}
\paragraph{The proposed architectural interventions effectively improve the verification and refinement of natural language explanations.}
Table \ref{tab:quantitative-explanation} and Figure \ref{fig:overall_results_explanation_syntax_error} compares our proposed framework with Explanation-Refiner on the tasks of verifying and refining natural language explanations across multiple LLMs. The results show that our approach more effectively and efficiently refines explanatory sentences for explanation-based NLI. In contrast, Explanation-Refiner achieves substantially lower refinement rates, for example, 51\% versus 78\% in e-SNLI for Llama3.1, 69\% versus 95\% for Deepseek-V3, 30\% versus 77\% for GPT-4o-mini, and 71\% versus 89\% for GPT-4o. Furthermore, Explanation-Refiner generally requires more iterations to refine each explanation, indicating that although it may identify specific logical errors, it is less efficient. For instance, Explanation-Refiner requires an average of 4.31 iterations in the QASC dataset, compared to 3.0 for our approach. Its performance is particularly limited on the WorldTree dataset, which contains complex, real-world scientific explanations requiring multi-hop reasoning. By contrast, our framework refines a significantly larger number of explanations in WorldTree, underscoring its capacity to handle more challenging inference scenarios. Furthermore, the current approach requires fewer LLM calls on average to fully refine an explanation, resulting in reduced inference time and cost compared to the Explanation-Refiner (See Table~\ref{tab:quantitative-explanation}). The average number of LLM calls was reduced by 58.60\%, 39.39\%, and 31.15\% on e-SNLI, QASC, and WorldTree, respectively, across all LLMs. Applying our approach with Deepseek-V3 on the e-SNLI dataset yields the most significant reduction, at 65.68\%.

\paragraph{The refinement process effectively corrects autoformalisation errors.} Figure \ref{fig:sub2_syntax_esnli}, \ref{fig:sub2_syntax_qasc}, and \ref{fig:sub2_syntax_worldtree} present the number of theories in the last iteration containing syntactic and inconsistency errors over five syntax error refinement iterations, comparing our proposed framework with Explanation-Refiner. Overall, our framework yields fewer syntactic errors. By incorporating syntactic parsing into autoformalisation, it guides LLMs to capture fine-grained logical properties of natural language sentences, thereby reducing type unification errors in constructed theories. Empirically, most syntactic errors diminish considerably within the first three iterations, after which the rate of improvement stabilises. The evaluation results of the number of theories that contain logical consistency errors are shown in Figure \ref{fig:_logical_inconsistency}.

\paragraph{Syntactic parsing improves faithfulness in autoformalisation.}
We convert the autoformalised logical forms back into natural language sentences using a rule-based algorithm that reconstructs each sentence from its action/verb predicates and corresponding argument information. We then calculate the cosine similarity between these reconstructed (informalised) sentences between the original sentences as the faithfulness of autoformalisation, as shown in Figure \ref{fig:faithfulness_utility}. Our approach shows a generally higher faithfulness compared to Explanation-Refiner, with an average of 0.7938, 0.7804, and 0.5975, compared to 0.6706, 0.5714, and 0.4220 across all three datasets. Our findings indicate that certain models exhibit comparatively lower similarity scores than others. Further investigation reveals that models such as Llama3.1-70b tend to generate non-existent predicates during formalisation in Explanation-Refiner, resulting in over-generation that undermines faithfulness and introduces extraneous information into the theory. More details about the rule-based algorithm are included in the Appendix \ref{rule-based}.

\paragraph{Logically guided proof sketches provide effective feedback for explanation refinement.} By constructing proof steps from logical propositions, relations, and derived implications, our method more precisely pinpoints logical errors, enabling the LLM to iteratively refine explanatory sentences in subsequent attempts to prove the theorem. As shown in Figure~\ref{fig:faithfulness_utility}, the average utility defined as the proportion of newly introduced explanations that are applied in the next iteration’s proof remains consistently higher for our approach compared to Explanation-Refiner, even as the number of iterations increases. In contrast, Explanation-Refiner’s utility markedly decreases over successive iterations.

\subsection{Ablation Study}
We conducted several ablation studies to evaluate the impact of the proposed components. Table \ref{tab:ablation_4o_deepseek} shows the results on GPT-4o and Deepseek-V3, while Table \ref{tab:ablation_manual_evaluation_full} in Appendix \ref{ablation_appendix} shows the full ablations.

\paragraph{Detailed feedback and syntax error refinement have the highest impact.} The most significant drop in performance is observed from removing detailed feedback and syntax error refinement steps. Providing detailed, step-level feedback to the LLM proves significantly more effective than using only a binary signal (i.e., \emph{provable} or \emph{unprovable}). When replacing detailed with binary feedback, the number of refined explanations dropped substantially; for instance, GPT-4o showed a 39\% decrease in refined explanations in the WorldTree dataset. Excluding the syntactic error refinement stage frequently yielded theories that failed under theorem prover scrutiny, thereby producing little to no useful feedback for subsequent refinement.

\paragraph{Logical expression aids LLMs in proof sketches generation, reducing hallucinations that could lead to incorrect or failed proof construction for explanation refinement.}
Eliminating the logical expression-guided proof step generation component led to an increase in required iterations for explanation refinement and a reduction in the total number of successfully refined explanations. These findings highlight the importance of logical expressions in constructing coherent proofs and mitigating hallucinations that otherwise result in incorrect or failed proofs.

% \paragraph{Variable and quantifier errors in the autoformalisation process significantly impact faithfulness. }
% We further conducted a human evaluation on three types of errors: variable errors (identifiable by the theorem prover), implication errors, and quantifier errors (not identifiable by the theorem prover) as shown in Table \ref{tab:ablation_4o_deepseek}. Our findings suggest that using LLMs for autoformalisation still leaves notable gaps, particularly in accurately handling variables and quantifiers. Although the number of explanations refined merely changed when we removed the quantifier refinement as shown in Table \ref{tab:ablation_4o_deepseek}, the number of quantifier errors increased from human evaluation. Explanation-Refiner does not apply a syntactic parsing and quantifier refinement resulting in more errors being introduced for variable, implication and quantifier errors as shown in Table \ref{tab:ablation_4o_deepseek}. Thus, we introduced both syntax error refinement and quantifier error refinement processes. Our results show a significant reduction in the overall error rate following the corresponding soft-critique model refinements.

\paragraph{Variable and quantifier errors significantly impact the faithfulness of autoformalisation. }
We further conducted a human evaluation on three types of errors: variable errors (identifiable by the theorem prover), implication errors, and quantifier errors (not identifiable by the theorem prover) as shown in Table \ref{tab:ablation_4o_deepseek}. Our findings suggest that using LLMs for autoformalisation still leaves notable gaps, particularly in accurately handling variables and quantifiers. As shown in Table \ref{tab:ablation_4o_deepseek}, removing the quantifier refinement did not substantially alter the number of refined explanations. However, human evaluation indicates that the number of quantifier-related errors increased when this refinement was omitted. Explanation-Refiner does not apply a syntactic parsing and quantifier refinement, resulting in more errors being introduced for variable, implication and quantifier errors as shown in Table \ref{tab:ablation_4o_deepseek}. Thus, we introduced both syntax error refinement and quantifier error refinement processes. Our results show a significant reduction in the overall error rate following the corresponding soft-critique model refinements.

\section{Related Work} 
\paragraph{LLM-Symbolic Integration} Recent approaches have attempted to integrate LLMs with external symbolic models through autoformalisation -- i.e., the translation of informal statements into formal representations. Recent work explores this task in both mathematical \citep{NEURIPS2022_d0c6bc64, Jiang2022DraftSA, agrawal2022mathematicsformalisationassistantusing, zhang-etal-2024-consistent, xin2024deepseekproveradvancingtheoremproving, lu2024formalalignautomatedalignmentevaluation, lu2024processdrivenautoformalizationlean4, leang2025theoremproverjudgesynthetic} and logical \citep{olausson-etal-2023-linc, quan-etal-2024-enhancing, kirtania-etal-2024-logic, lee2025entailmentpreservingfirstorderlogicrepresentations, raza2025instantiationbasedformalizationlogicalreasoning} domains using the support of automated theorem provers. Several studies \citep{pan-etal-2023-logic, jiang-etal-2024-leanreasoner, quan-etal-2024-verification} transform natural language sentences into logical forms based on LLMs. \citet{qi2025largelanguagemodelsmeet} integrate symbolic provers into the generation process of logical reasoning problems and propose a multi-hop first-order logic (FOL) reasoning dataset. In contrast, our work tackles real-world occurrences of material inferences rather than purely synthetic data, thereby requiring more robust semantic representations and autoformalisation process to capture the complexity of multi-step reasoning over material inferences.

\paragraph{Theorem-Proving with LLMs} Proof generation refers to the task of generating intermediate proof steps as tactic predictions in automated theorem proving \citep{li2024dl4tp}. Recent work harness LLMs to produce formal proof scripts \citep{polu2020generativelanguagemodelingautomated, 51693,zhao2023decomposingenigmasubgoalbaseddemonstration, xin2024deepseekproverv15harnessingproofassistant, first2023baldurwholeproofgenerationrepair,frieder2024largelanguagemodelsmathematicians,welleck2023llmstep, welleck2023llmstepllmproofstepsuggestions, thakur2024incontextlearningagentformal, 10.1145/3626252.3630928}, often by translating high-level reasoning into low-level tactics. \citet{quan-etal-2024-verification}, for example, prompts the LLM to first produce a rough, informal inference strategy in natural language, and then automatically formalise this strategy into Isabelle/HOL proof steps. \citet{jiang-etal-2024-leanreasoner} utilises a tactic generator and a proof search module to select tactics during proof construction to build a proof tree from the root goal. \citet{liu2025logicofthoughtinjectinglogiccontexts} use propositional logic, introduce Logic-of-Thought prompting, a technique that appends logically enriched descriptions to the original context, thereby improving informational completeness and bolstering the LLMs logical reasoning ability. In contrast, our approach synthesises logical reasoning guidance in close iterative dialogue with automated provers to provide more robust and interpretable proofs in contrast to LLM-driven single-pass methods.

\section{Conclusion}
% In this paper, we proposed formally-guided methods to address the challenges involved in using external theorem provers to verify and refine natural language explanations in the domain of natural language inference. By incorporating syntactic parsing, targeted syntactic error checking, logical-relation guidance, and detailed feedback at each proof step, our approach significantly outperforms prior work in both faithfulness of autoformalisation and robustness of iterative explanation refinement. Ablation studies highlight the critical role of these components in reducing syntactic errors, maintaining consistency, and promoting more efficient logical verification and refinement.

In this paper, we proposed formally-guided methods to address the challenges involved in using external theorem provers to verify and refine natural language explanations in the domain of natural language inference. By incorporating syntactic parsing, targeted syntactic error checking, logical-relation guidance, and detailed feedback at each proof step, our approach significantly outperforms prior work in both faithfulness of autoformalisation and robustness of iterative explanation refinement. Ablation studies underscore the importance of each component in reducing syntactic issues, maintaining consistency, and promoting more efficient logical verification and refinement. This framework opens avenues for more transparent, reliable, and scalable NLI systems. Moving forward, we plan to explore more advanced theorem-proving strategies with LLMs and domain-specific expansions, ultimately advancing toward increasingly interpretable and robust end-to-end NLI pipelines.

\section*{Limitations} Although our framework substantially improves both the consistency of autoformalisation and the robustness of explanation verification, certain limitations remain. First, LLMs can still introduce variable inconsistencies, erroneous implications, and incorrect quantifiers that are not fully resolved by automated checking. Second, some explanations require nuanced real-world knowledge or domain-specific axioms that exceed current formal reasoning capabilities, requiring expert oversight. Finally, the reliability of our iterative refinement pipeline hinges on high-quality LLM output and proof-step feedback; degraded model performance or noisy system responses can hinder successful verification. Future work may explore more advanced semantic checks, stronger model calibration, and selective human intervention to further enhance faithfulness and correctness.

\section*{Ethical statement} While this work focuses on the introduction of mechanisms for improving the control and logical consistency properties of LLM-based NLI, having an overall positive impact, further investigations are needed to understand the specific conditions in which these methods can perform. The application of these methods on real-world or critical settings need to be complemented by human supervision or extensive quantitative and qualitative assessment.

\section*{Acknowledgments}
This work was partially funded by the SNSF project NeuMath (\href{https://data.snf.ch/grants/grant/204617}{200021\_204617}), by the EPSRC grant EP/T026995/1, “EnnCore” under Security for all in an AI-enabled society, by the CRUK National Biomarker Centre, and supported by the Manchester Experimental Cancer Medicine Centre and the NIHR Manchester Biomedical Research Centre.

\bibliography{anthology,custom}
%\bibliography{custom}

\appendix
\label{sec:logical_inconsistencies}
\begin{figure*}[t]
\centering
\begin{subfigure}{.3\textwidth} 
  \centering
  \includegraphics[width=\linewidth]{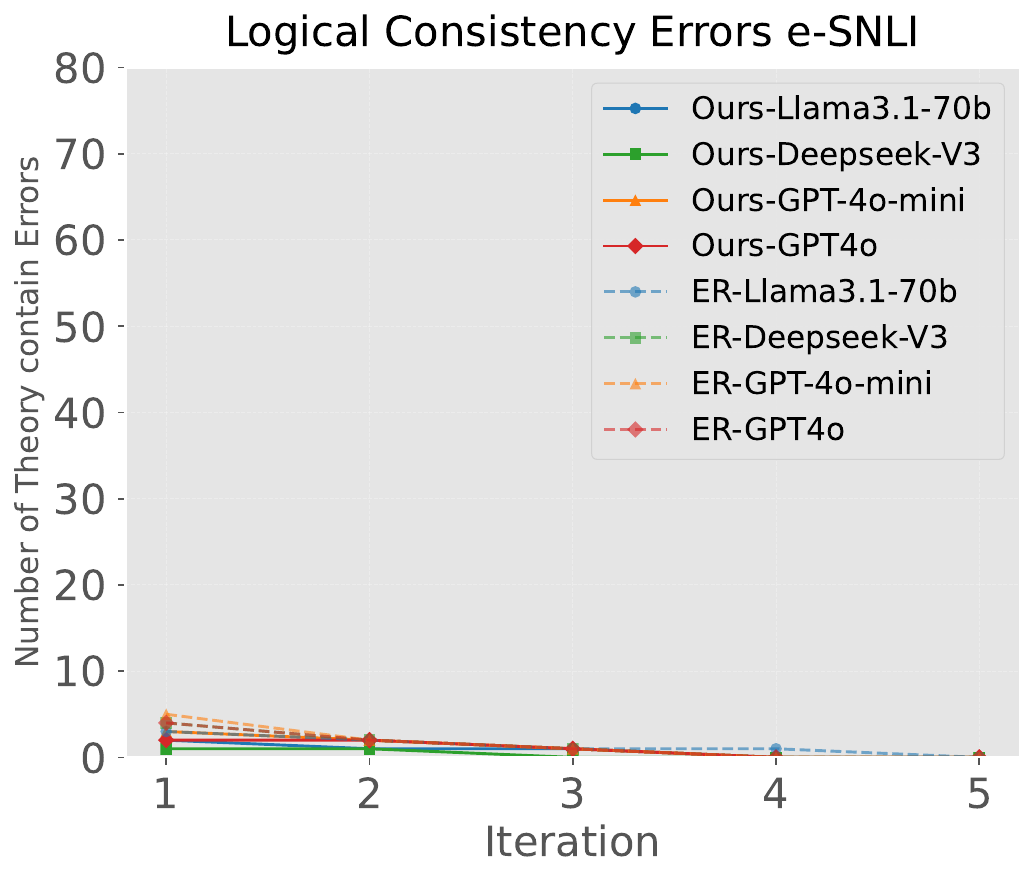}
  \caption{}
  \label{fig:sub1_syntax}
\end{subfigure}
\hfill
\begin{subfigure}{.3\textwidth} 
  \centering
  \includegraphics[width=\linewidth]{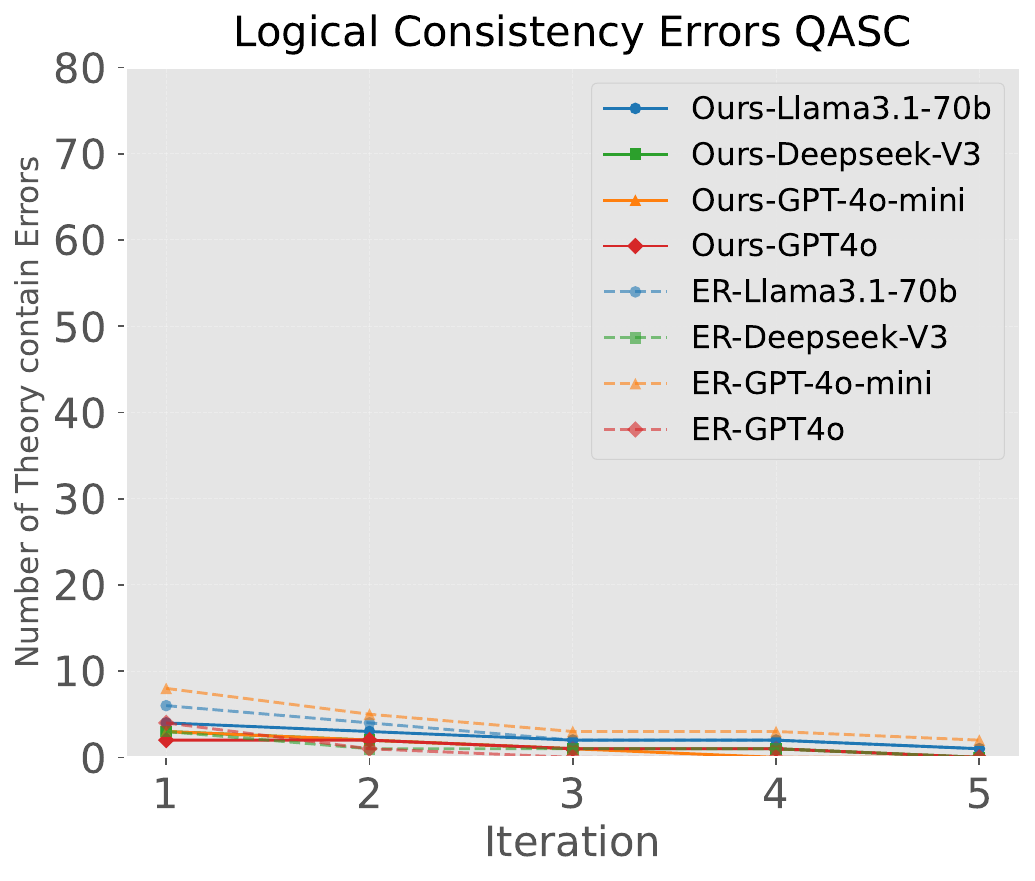}
  \caption{}
  \label{fig:sub2_syntax}
\end{subfigure}
\hfill
\begin{subfigure}{.3\textwidth} 
  \centering
  \includegraphics[width=\linewidth]{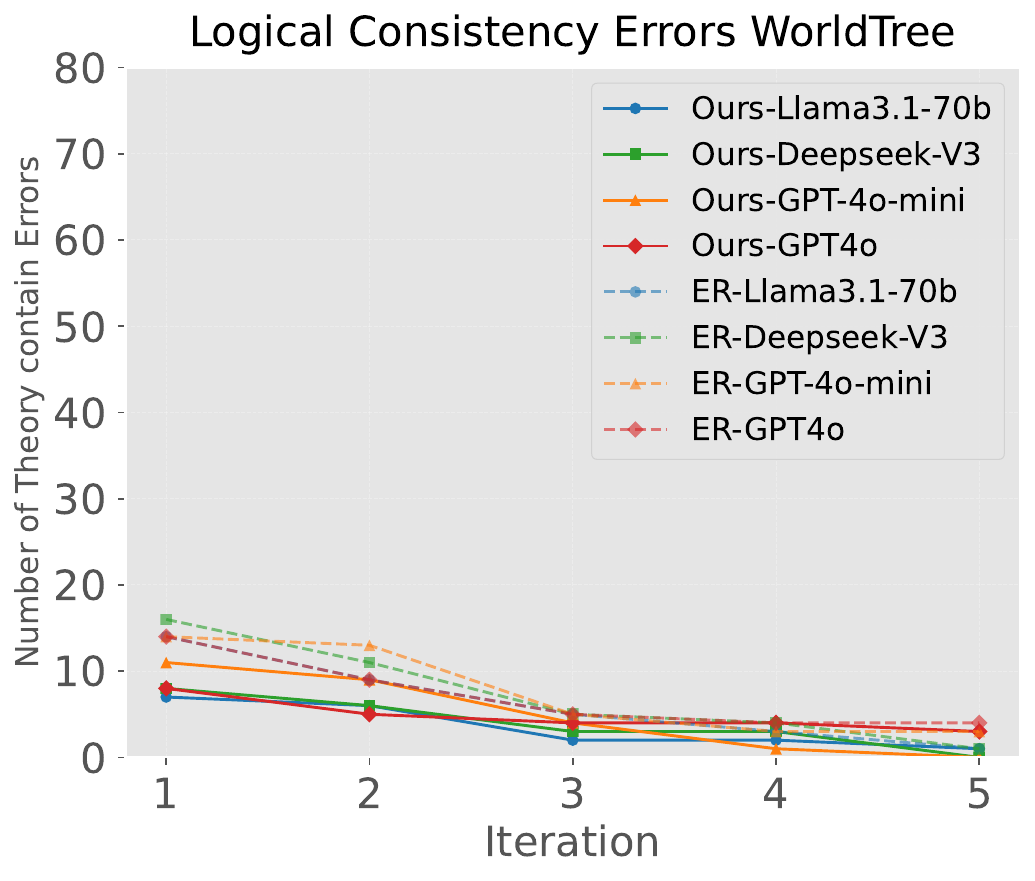}
  \caption{}
\end{subfigure}
\caption{Number of theories that contain logical consistency error at each syntax error refinement stage.}
\label{fig:_logical_inconsistency}
\end{figure*}

\section{Ablation study}
Table \ref{tab:ablation_manual_evaluation_full} shows the overall results on the ablation study for all LLMs.
\label{ablation_appendix}
\begin{table*}[t]
\centering
\tiny
\renewcommand{\arraystretch}{1.15}
\setlength{\tabcolsep}{4pt}
\sisetup{number-unit-product = {}}
\begin{tabular}{
@{}l
  S[table-format=3.2]   % e-SNLI Init.
  S[table-format=4.2]   % e-SNLI Final. 
  S[table-format=1.2]   % e-SNLI #Iter
  S[table-format=3.2]   % QASC Init.
  S[table-format=4.2]   % QASC Final.
  S[table-format=1.2]   % QASC #Iter
  S[table-format=2.2]   % WorldTree Init.
  S[table-format=4.2]   % WorldTree Final.
  S[table-format=1.2]   % WorldTree #Iter
@{}}
\toprule
& \multicolumn{3}{c}{\textbf{e-SNLI}}
& \multicolumn{3}{c}{\textbf{QASC}}
& \multicolumn{3}{c}{\textbf{WorldTree}} \\
\cmidrule(lr){2-4}\cmidrule(lr){5-7}\cmidrule(lr){8-10}
\textbf{}   
& \textbf{Init.} & \textbf{Final} & \textbf{\#Iter}
& \textbf{Init.} & \textbf{Final} & \textbf{\#Iter}
& \textbf{Init.} & \textbf{Final} & \textbf{\#Iter} \\
\midrule
\multicolumn{10}{@{}l}{\textit{\textbf{Ablations on our approach}}} \\
\midrule
Llama3.1-70b  (- logical relations)
 & \SI{34}{\percent}  & \SI{74}{\percent(-4\percent)} & 2.43
 & \SI{9}{\percent} & \SI{58}{\percent(-10\percent)} & 2.94
 & \SI{7}{\percent} & \SI{44}{\percent(-8\percent)} & 5.42 \\
Llama3.1-70b  (- \textbf{detailed feedback})
 & \SI{32}{\percent} & \SI{66}{\percent(-12\percent)} & 3.42
 & \SI{10}{\percent} & \SI{34}{\percent(-\textbf{\underline{34}}\percent)} & 3.64
 & \SI{5}{\percent} & \SI{24}{\percent(-\textbf{\underline{28}}\percent)} & 8.12 \\
 Llama3.1-70b  (- refine quantifiers)
 & \SI{28}{\percent} & \SI{77}{\percent(-1\percent)} & 2.18
 & \SI{9}{\percent} & \SI{68}{\percent(-0\percent)} & 2.88
 & \SI{3}{\percent} & \SI{50}{\percent(-2\percent)} & 4.52 \\
 Llama3.1-70b  (- \textbf{refine syntax errors})
 & \SI{18}{\percent} & \SI{57}{\percent(-\textbf{\underline{21}}\percent)} & 4.58
 & \SI{5}{\percent} & \SI{53}{\percent(-15\percent)} & 4.47
 & \SI{3}{\percent} & \SI{30}{\percent(-20\percent)} & 6.12 \\
\hdashline[3pt/4pt]  
GPT-4o-mini (- logical relations)
 & \SI{27}{\percent}  & \SI{65}{\percent(-12\percent)} & 2.31
 & \SI{11}{\percent} & \SI{57}{\percent(-14\percent)} & 4.12
 & \SI{6}{\percent} & \SI{27}{\percent(-20\percent)} & 5.19 \\
GPT-4o-mini (- detailed feedback)
 & \SI{30}{\percent} & \SI{62}{\percent(-15\percent)} & 4.56
 & \SI{9}{\percent} & \SI{46}{\percent(-25\percent)} & 3.87
 & \SI{3}{\percent} & \SI{19}{\percent(-28\percent)} & 6.21 \\
GPT-4o-mini  (- refine quantifiers)
 & \SI{26}{\percent} & \SI{78}{\percent(+4\percent)} & 2.10
 & \SI{5}{\percent} & \SI{73}{\percent(+2\percent)} & 2.92
 & \SI{4}{\percent} & \SI{46}{\percent(-1\percent)} & 5.13 \\
GPT-4o-mini  (- \textbf{refine syntax errors})
 & \SI{15}{\percent} & \SI{43}{\percent(-\textbf{\underline{34}}\percent)} & 2.86
 & \SI{3}{\percent} & \SI{34}{\percent(-\textbf{\underline{37}}\percent)} & 3.65
 & \SI{3}{\percent} & \SI{10}{\percent(-\textbf{\underline{37}}\percent)} &5.21 \\
\hdashline[3pt/4pt]  
GPT-4o (- \textbf{logical relations})
 & \SI{34}{\percent} & \SI{74}{\percent(-\textbf{\underline{15}}\percent)} & 2.24
 & \SI{12}{\percent} & \SI{58}{\percent(-21\percent)} & 3.46
 & \SI{6}{\percent} & \SI{38}{\percent(-18\percent)} & 4.36 \\
GPT-4o (- \textbf{detailed feedback})
 & \SI{35}{\percent} & \SI{83}{\percent(-6\percent)} & 2.86
 & \SI{13}{\percent} & \SI{56}{\percent(-\textbf{\underline{23}}\percent)} & 4.45
 & \SI{5}{\percent} & \SI{17}{\percent(-\textbf{\underline{39}}\percent)} & 6.46 \\
GPT-4o  (- refine quantifiers)
 & \SI{34}{\percent} & \SI{87}{\percent(-2\percent)} & 1.63
 & \SI{14}{\percent} & \SI{83}{\percent(+4\percent)} & 2.89
 & \SI{7}{\percent} & \SI{49}{\percent(-7\percent)} & 3.65 \\
GPT-4o  (- \textbf{refine syntax errors})
 & \SI{21}{\percent} & \SI{74}{\percent(-\textbf{\underline{15}}\percent)} & 2.34
 & \SI{5}{\percent} & \SI{58}{\percent(-21\percent)} & 4.11
 & \SI{2}{\percent} & \SI{24}{\percent(-32\percent)} &6.48 \\
\hdashline[3pt/4pt]
Deepseek-V3  (- logical relations)
 & \SI{39}{\percent}  & \SI{89}{\percent(-6\percent)} & 1.68
 & \SI{16}{\percent} & \SI{77}{\percent(-13\percent)} & 2.64
 & \SI{10}{\percent} & \SI{58}{\percent(-15\percent)} & 4.01 \\
Deepseek-V3  (- \textbf{detailed feedback})
 & \SI{31}{\percent} & \SI{86}{\percent(-9\percent)} & 3.22
 & \SI{22}{\percent} & \SI{69}{\percent(-21\percent)} & 4.12
 & \SI{6}{\percent} & \SI{41}{\percent(-\textbf{\underline{32}}\percent)} & 6.13 \\
Deepseek-V3  (- refine quantifiers)
 & \SI{34}{\percent} & \SI{96}{\percent(+1\percent)} & 1.64
 & \SI{14}{\percent} & \SI{93}{\percent(+3\percent)} & 1.89
 & \SI{6}{\percent} & \SI{70}{\percent(-3\percent)} & 3.23 \\
Deepseek-V3  (- \textbf{refine syntax errors})
 & \SI{28}{\percent} & \SI{77}{\percent(-\textbf{\underline{18}}\percent)} & 2.69
 & \SI{12}{\percent} & \SI{68}{\percent(-\textbf{\underline{22}}\percent)} & 2.84
 & \SI{4}{\percent} & \SI{46}{\percent(-27\percent)} &5.32 \\
 \midrule
\end{tabular}

\begin{tabular}{
@{}l
  S[table-format=3.2]   
  S[table-format=4.2]   
  S[table-format=2.2]   
  S[table-format=3.2]   
  S[table-format=4.2]   
  S[table-format=2.2]   
  S[table-format=2.2]   
  S[table-format=4.2]   
  S[table-format=2.2]   
@{}}
\toprule
& \multicolumn{3}{c}{\textbf{e-SNLI}}
& \multicolumn{3}{c}{\textbf{QASC}}
& \multicolumn{3}{c}{\textbf{WorldTree}} \\
\cmidrule(lr){2-4}\cmidrule(lr){5-7}\cmidrule(lr){8-10}
\textbf{}   
& \textbf{V.} & \textbf{I.} & \textbf{Q.}
& \textbf{V.} & \textbf{I.} & \textbf{Q.}
& \textbf{V.} & \textbf{I.} & \textbf{Q.} \\
\midrule
\multicolumn{10}{@{}l}{\textit{\textbf{Explanation-Refiner}}} \\
\midrule
Llama3.1-70b 
 & \SI{24}{\percent} & \SI{10}{\percent} & \SI{10}{\percent}
 & \SI{43}{\percent} & \SI{12}{\percent} & \SI{34}{\percent}
 & \SI{45}{\percent} & \SI{15}{\percent} & \SI{27}{\percent} \\
GPT-4o-mini
  & \SI{18}{\percent} & \SI{8}{\percent} & \SI{7}{\percent}
 & \SI{41}{\percent} & \SI{8}{\percent} & \SI{32}{\percent}
 & \SI{39}{\percent} & \SI{9}{\percent} & \SI{29}{\percent} \\
GPT-4o
  & \textbf{\underline{\SI{9}{\percent}}} & \textbf{\underline{\SI{3}{\percent}}} & \textbf{\underline{\SI{6}{\percent}}}
 & \textbf{\underline{\SI{18}{\percent}}} & \textbf{\underline{\SI{9}{\percent}}} & \textbf{\underline{\SI{18}{\percent}}}
 & \textbf{\underline{\SI{33}{\percent}}} & \textbf{\underline{\SI{8}{\percent}}} & \textbf{\underline{\SI{16}{\percent}}} \\
Deepseek-V3 
  & \SI{27}{\percent} & \textbf{\underline{\SI{3}{\percent}}} & \SI{10}{\percent}
 & \SI{34}{\percent} & \SI{9}{\percent} & \SI{25}{\percent}
 & \SI{44}{\percent} & \SI{23}{\percent} & \SI{31}{\percent} \\
\midrule
\multicolumn{10}{@{}l}{\textit{\textbf{Ablations}}} \\
\midrule
 Llama3.1-70b  (- refine quantifiers)
 & \SI{23}{\percent} & \SI{8}{\percent} & \SI{13}{\percent(+3\percent)}
 & \SI{39}{\percent} & \SI{11}{\percent} & \SI{43}{\percent(+9\percent)}
 & \SI{43}{\percent} & \SI{13}{\percent} & \SI{36}{\percent(+9\percent)} \\
 Llama3.1-70b  (- refine syntax errors)
 & \SI{41}{\percent(+17\percent)} & \SI{10}{\percent} & \SI{9}{\percent}
 & \SI{53}{\percent(+10\percent)} & \SI{8}{\percent} & \SI{34}{\percent}
 & \SI{65}{\percent(+20\percent)} & \SI{11}{\percent} & \SI{23}{\percent} \\
\hdashline[3pt/4pt]
GPT-4o-mini  (- refine quantifiers)
  & \SI{14}{\percent} & \SI{5}{\percent} & \SI{11}{\percent(+4\percent)}
 & \SI{35}{\percent} & \SI{10}{\percent} & \SI{47}{\percent(+15\percent)}
 & \SI{41}{\percent} & \SI{10}{\percent} & \SI{34}{\percent(+5\percent)} \\
GPT-4o-mini  (- refine syntax errors)
 & \SI{39}{\percent(+21\percent)} & \SI{4}{\percent} & \SI{9}{\percent}
 & \SI{63}{\percent(+21 \percent)} & \SI{7}{\percent} & \SI{12}{\percent}
 & \SI{55}{\percent(+16\percent)} & \SI{7}{\percent} & \SI{23}{\percent} \\
\hdashline[3pt/4pt]  
GPT-4o  (- refine quantifiers)
  & \SI{9}{\percent} & \SI{2}{\percent} & \SI{10}{\percent(+4\percent)}
 & \SI{13}{\percent} & \SI{2}{\percent} & \SI{23}{\percent(+5\percent)}
 & \SI{38}{\percent} & \SI{6}{\percent} & \SI{19}{\percent(+3\percent)} \\
GPT-4o  (- refine syntax errors)
 & \SI{16}{\percent(+7\percent)} & \SI{1}{\percent} & \SI{3}{\percent}
 & \SI{38}{\percent(+20\percent)} & \SI{3}{\percent} & \SI{7}{\percent}
 & \SI{56}{\percent(+23\percent)} & \SI{3}{\percent} & \SI{16}{\percent} \\
\hdashline[3pt/4pt]  
Deepseek-V3  (- refine quantifiers)
 & \SI{25}{\percent} & \SI{5}{\percent} & \SI{16}{\percent(+6\percent)}
 & \SI{31}{\percent} & \SI{14}{\percent} & \SI{35}{\percent(+10\percent)}
 & \SI{32}{\percent} & \SI{11}{\percent} & \SI{38}{\percent(+7\percent)} \\
Deepseek-V3  (- refine syntax errors)
 & \SI{38}{\percent(+11\percent)} & \SI{4}{\percent} & \SI{11}{\percent}
 & \SI{51}{\percent(+17\percent)} & \SI{9}{\percent} & \SI{15}{\percent}
 & \SI{67}{\percent(+23\percent)} & \SI{21}{\percent} & \SI{27}{\percent} \\
\bottomrule
\end{tabular}
\caption{Top -- Ablation study on the impacts of removing components on the analysis of number of explanation refined across three datasets. Bottom -- Comparison of manually evaluated variable, implication, and quantifier errors in the autoformalisation process from a randomly sampled set of 100 Isabelle/HOL theories across all iterations for each LLM.}
\label{tab:ablation_manual_evaluation_full}
\end{table*}

\begin{figure*}
\begin{lstlisting}
Original Sentence: The boy is inside of the building.
Logical Form 1: $\exists$x y e. Boy(x) $\wedge$ Building(y) $\wedge$ Inside(e) $\wedge$ Agent(e, x) $\wedge$ Patient(e, y)
Informalised Sentence 1: Boy in side building.
Sentence Similarity: 0.9344

Logical Form 2: $\exists$x y e. Boy(x) $\wedge$ Building(y) $\wedge$ Inside(e) $\wedge$ Agent(e, x)
Informalised Sentence 2: Boy in side.
Sentence Similarity: 0.8127
\end{lstlisting}
\caption{An example of the faithfulness between two informalised logical forms}
\label{fig:rule_based}
\end{figure*}

\section{Informalisation}
\label{rule-based}
We perform an autoformalisation process that transforms natural language sentences into Neo-Davidsonian event-based semantics by leveraging their underlying structure. One way to measure the faithfulness of this autoformalisation is to translate the constructed logical forms back into natural language and then compare the generated (informalised) sentences with the original ones using cosine similarity.

We employ a rule-based method to transform Neo-Davidsonian logical forms back into coherent natural language. First, we parse a logical form that may contain multiple conjuncts, typically connected by the logical “and” operator (\(\land\)). Each conjunct is treated as an atomic predicate with the general structure \(\text{Predicate}(\text{arg}_1, \text{arg}_2, \dots)\). Once the form is separated into atomic predicates, we distinguish between role predicates (e.g., \(\texttt{Agent}(e_1,x)\), \(\texttt{Patient}(e_1,y)\)) and entity-attribute predicates (e.g., \(\texttt{Child}(x)\), \(\texttt{Blonde}(x)\)). The role predicates specify how each entity participates in the event (agent, patient, theme, location, etc.), while the attribute predicates detail intrinsic properties of those entities (for instance, “child,” “blonde,” “small,” or “plastic”).

After identifying these predicates, we group together all attributes describing the same entity variable. In particular, we parse the attributes from right to left, treating the rightmost attribute as the head noun and the preceding ones as adjectives. For example, if a single entity \(x\) is associated with \(\texttt{Child}(x)\) and \(\texttt{Blonde}(x)\), we combine those attributes to form a concise descriptor such as “blonde child.” Likewise, if another entity \(y\) has attributes \(\texttt{Plastic}(y)\) and  \(\texttt{Small}(y)\), we might call it “small plastic”.

Next, we convert these role–entity pairings into simple event-level sentences. For each event \(e_i\), we identify which entity is the \(\texttt{Agent}\) and which is the \(\texttt{Patient}\) (or any other role labels), then build a straightforward sentence. For instance, if \(x\) is “blonde child” and \(y\) is “small plastic item,” the corresponding natural language description might by constructed from the event verb “Puts” as “blonde child puts small plastic item” The specific event verb (“puts,” “picks,” “hands over,” etc.) would depend on how the event predicate itself is represented in the logical form.

In cases where the logical form contains implication, we divide the logical forms into sub-formulas. Complex operators and connectives (i.e. \(\lor\)) will be mapped carefully to their closest equivalents in English.

As shown in Figure \ref{fig:rule_based}, different formalised logical forms can affect the faithfulness of the autoformalisation. For instance, Logical Form 2 omits the \textit{Patient} argument, causing the rule-based system to skip translating the predicate information for the \textit{building} back into natural language, and thus producing an unfaithful representation.

\begin{algorithm*}
\small
\DontPrintSemicolon
\SetKwInOut{Input}{Input}
\SetKwInOut{Output}{Output}
\SetKwFunction{ProcessLogicalProposition}{ProcessLogicalProposition}
\SetKwFunction{ParseInput}{ParseInput}

\Input{logical\_information: string with propositions and relations}
\Output{result: string with processed relations and implications}

logical\_props, logical\_exprs $\gets$ \ParseInput{logical\_information}\;
\BlankLine

\eIf{logical\_exprs = $\emptyset$}{
    result $\gets$ format\_propositions(logical\_props)\;
    \Return result\;
}{
    Initialise symbols\_dict, symbol\_meanings $\gets \{\}$\;
    \ForEach{(key, value) $\in$ logical\_props}{
        sanitized\_key $\gets$ sanitize(key)\;
        symbol $\gets$ create\_symbol(sanitized\_key)\;
        Update symbols\_dict and symbol\_meanings\;
    }
    
    \tcp{Define SymPy logical operators dictionary}
    logical\_operators $\gets \{$\;
    \Indp
        symbols\_dict,  
        Not: SymPy negation,\;
        And: SymPy conjunction,\;
        Or: SymPy disjunction,\;
        Implies: SymPy implication,\;
        Equivalent: SymPy equivalence\;
    \Indm$\}$\;
    
    propositions $\gets []$\;
    initial\_implications $\gets \emptyset$\;
    
    \ForEach{expr $\in$ logical\_exprs}{
        expr $\gets$ replace\_symbols(expr)\;
        \tcp{Evaluate using SymPy's logical operators}
        prop $\gets$ evaluate\_with\_sympy(expr, logical\_operators)\;
        \tcp{Apply SymPy's simplification rules}
        simplified\_prop $\gets$ sympy.simplify(prop)\;
        propositions.append(prop)\;
        initial\_implications.add(simplified\_prop)\;
    }
    
    derived\_implications $\gets \emptyset$\;
    logical\_atoms $\gets$ get\_atoms(propositions)\;
    literals $\gets$ logical\_atoms $\cup$ \{$\neg$atom $\mid$ atom $\in$ logical\_atoms\}\;
    
    \tcp{Use SymPy's satisfiability checker}
    \ForEach{(antecedent, consequent) $\in$ literals $\times$ literals}{
        \If{antecedent $\neq$ consequent}{
            implication $\gets$ antecedent $\implies$ consequent\;
            \tcp{Check using SymPy's logical rules}
            is\_new $\gets$ $\neg$equivalent\_to\_any(implication, initial\_implications)\;
            is\_valid $\gets$ check\_entailment(propositions, implication)\;
            \If{is\_new \textbf{and} is\_valid}{
                derived\_implications.add(implication)\;
            }
        }
    }
    
    result $\gets$ format\_output(logical\_props, logical\_exprs, derived\_implications)\;
    \Return result\;
}

\caption{Deriving logical implications with SymPy}
\label{algorithm_1}
\end{algorithm*}

\section{Datasets, LLMs and Theorem Prover}
The datasets used in our experiments are sourced from open academic works and include samples from e-SNLI \citep{NEURIPS2018_4c7a167b}, QASC \citep{Khot2019QASC}, and WorldTree \citep{jansen-etal-2018-worldtree}. We employed Isabelle/HOL \citep{nipkow2002isabelle} as the theorem prover, which is distributed under the revised BSD license, and used Explanation-Refiner \citep{quan-etal-2024-verification} as our baseline work, which is under the MIT license. Additionally, we utilised API calls for GPT-4o (gpt-4o-2024-08-06) \citep{DBLP:journals/corr/abs-2303-08774}, GPT-4o-mini (gpt-4o-mini-2024-07-18) \citep{DBLP:journals/corr/abs-2303-08774}, Deepseek-V3 (Deepseek-V3-671b) \citep{deepseekai2024deepseekv3technicalreport}, and Llama3.1-70b (LLama3.1-70b-Instruct) \citep{grattafiori2024llama3herdmodels}. All temperature is set to 0.

\section{Runtime Examples}
\begin{figure*}[t]
    \centering
    \includegraphics[width=\textwidth]{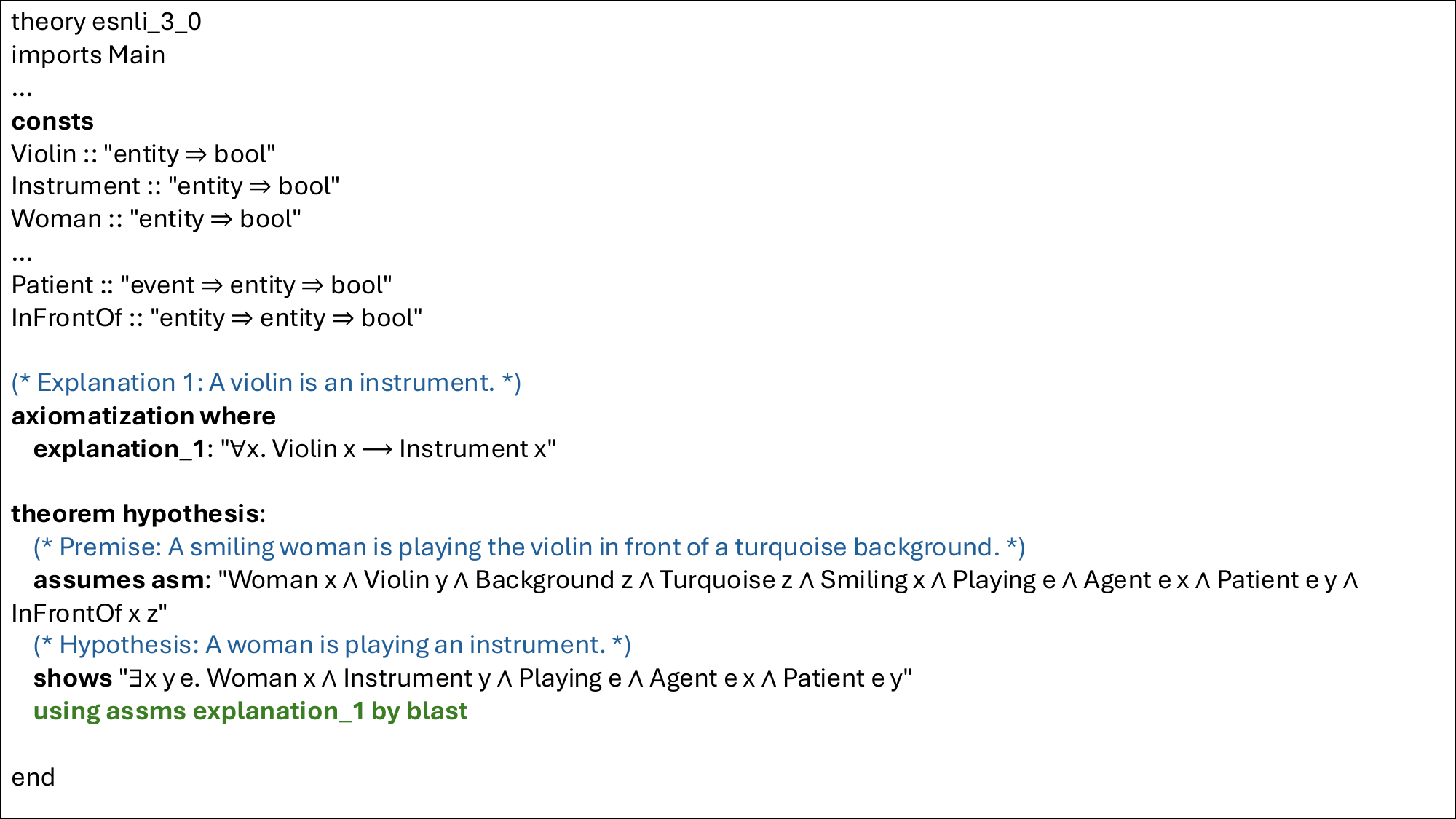}
    \caption{An example of an autoformalised Isabelle/HOL theory in the e-SNLI dataset.}
\label{fig:esnli-0}
\end{figure*}

\begin{figure*}[t]
    \centering
    \includegraphics[width=\textwidth]{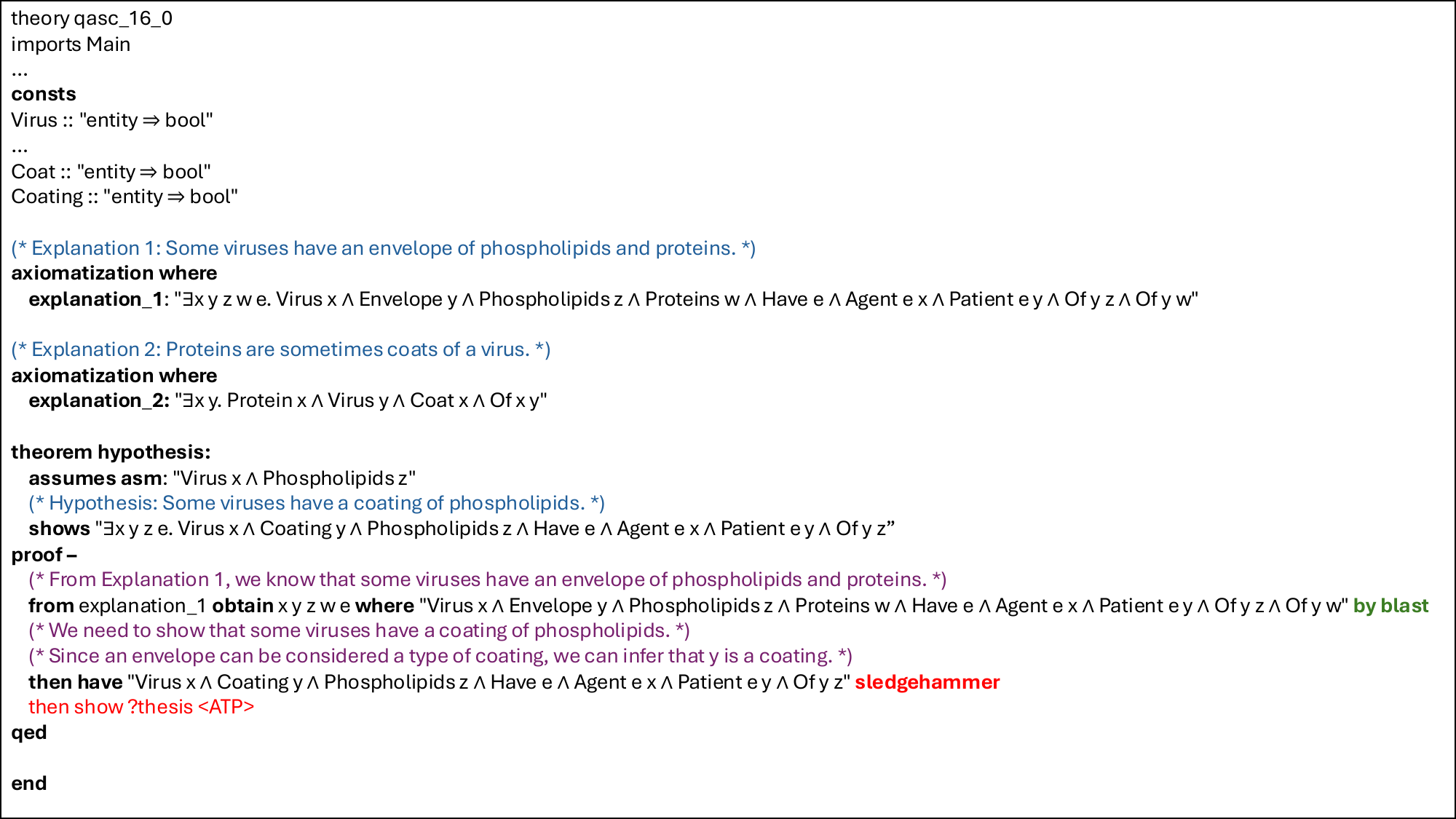}
    \caption{An example of an autoformalised Isabelle/HOL theory in the QASC dataset.}
\label{fig:qasc-0}
\end{figure*}

\begin{figure*}[t]
    \centering
    \includegraphics[width=\textwidth]{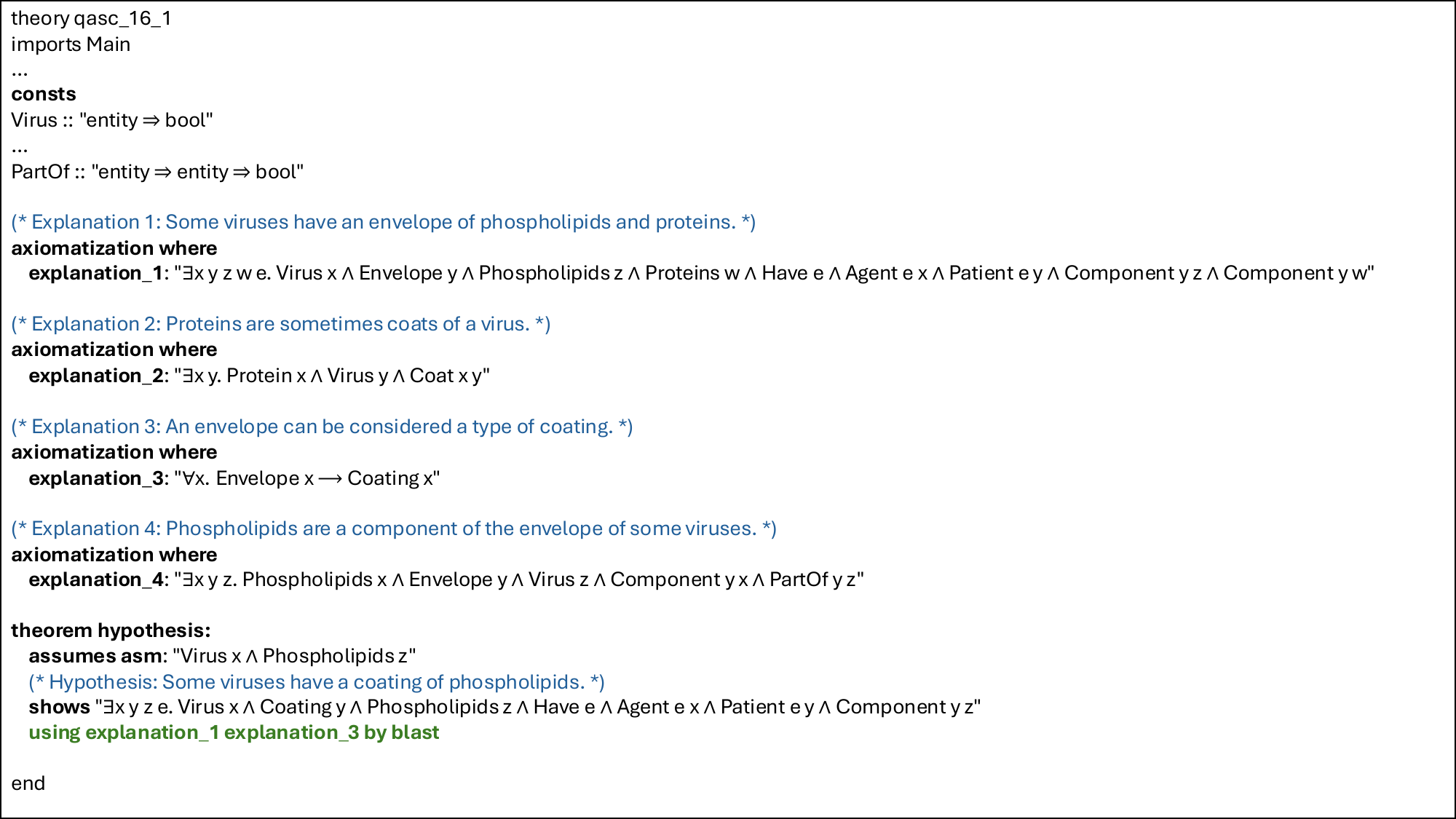}
    \caption{An example of an autoformalised Isabelle/HOL theory in the QASC dataset.}
\label{fig:qasc-1}
\end{figure*}

\begin{figure*}[t]
    \centering
    \includegraphics[width=\textwidth]{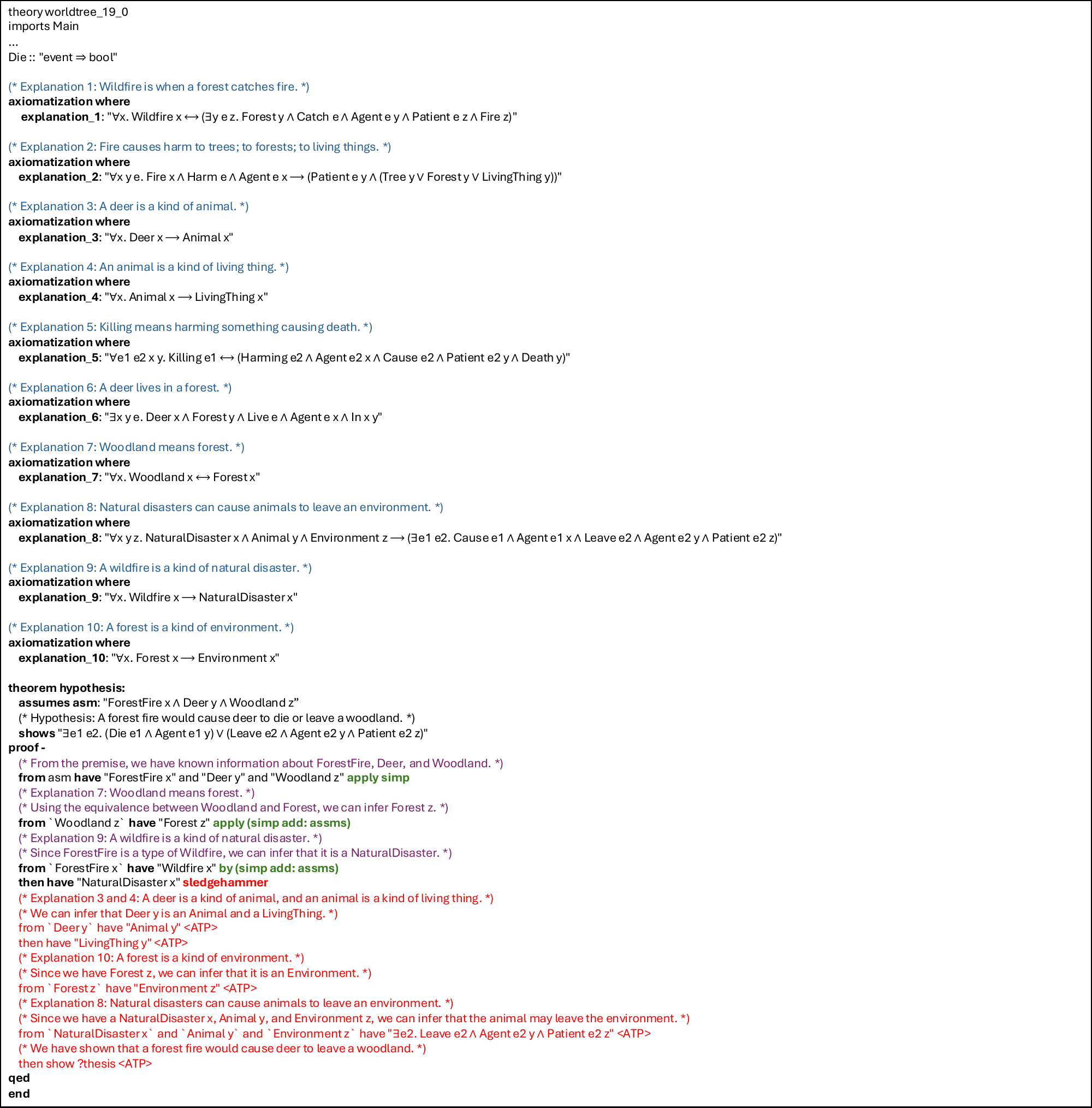}
    \caption{An example of an autoformalised Isabelle/HOL theory in the WorldTree dataset.}
\label{fig:wordltree-0}
\end{figure*}

\begin{figure*}[t]
    \centering
    \includegraphics[width=\textwidth]{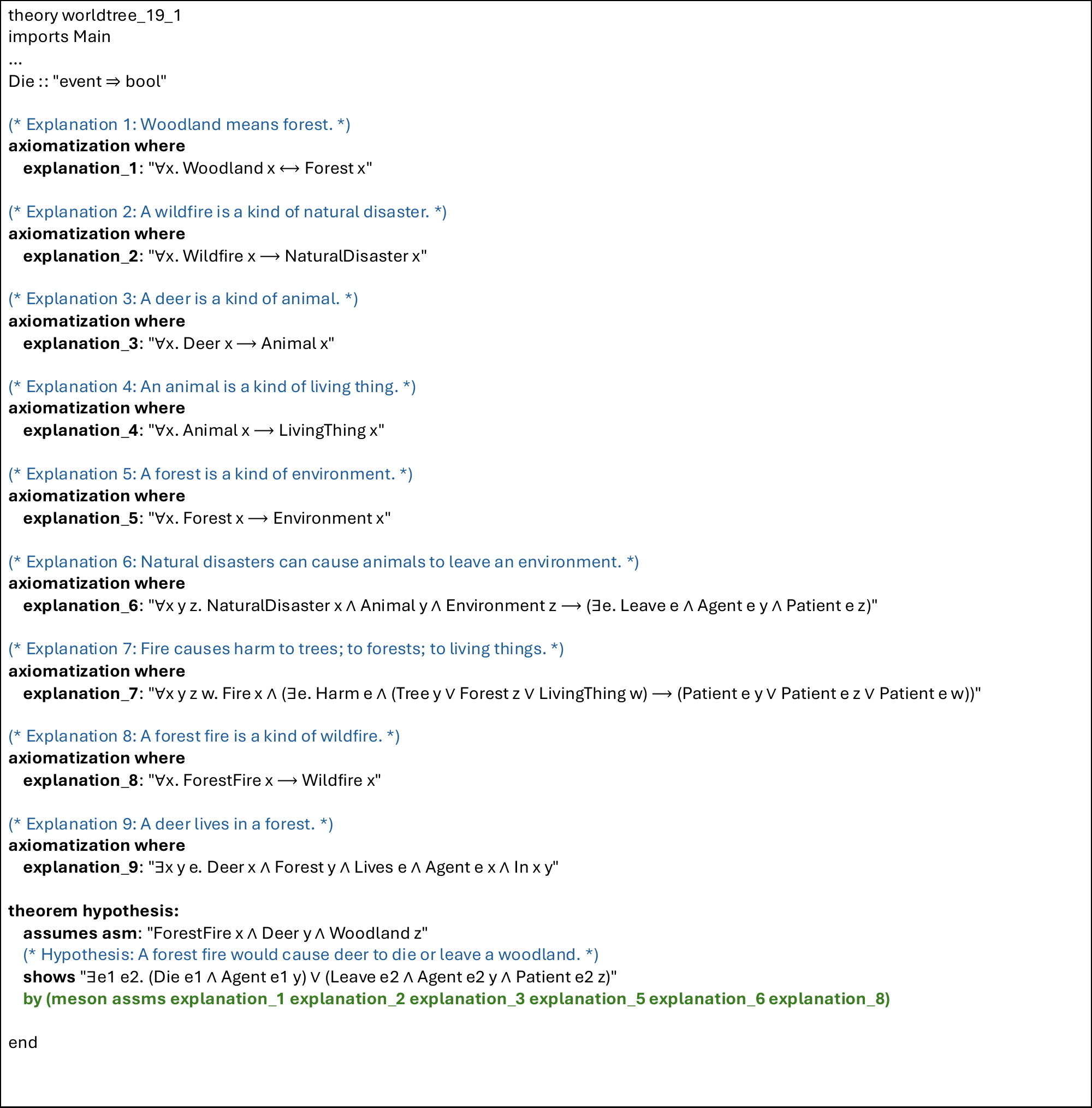}
    \caption{An example of an autoformalised Isabelle/HOL theory in the WorldTree dataset.}
\label{fig:worldtree-1}
\end{figure*}

Tables \ref{tab:e-snli_example_0_table}, \ref{tab:qasc_example_1_table}, and \ref{tab:worldtree_example_1_table}, together with Figures \ref{fig:esnli-0}, \ref{fig:qasc-0}, \ref{fig:qasc-1}, \ref{fig:wordltree-0}, and \ref{fig:worldtree-1}, show the runtime examples for the e-SNLI, QASC, and WorldTree datasets, respectively.

\begin{table*}[t]
\centering
\small
\begin{tabular}{@{}p{1cm}p{3.5cm}p{3cm}p{4cm}p{1cm}p{1cm}@{}}  
\toprule
\textbf{Dataset} & \textbf{Problem} & \textbf{Explanation} & \textbf{Logic Info} & \textbf{Iteration} & \textbf{Validity} \\
\midrule
e-SNLI & \textbf{Premise}: A smiling woman is playing the violin in front of a turquoise background. \newline \textbf{Hypothesis}: A woman is playing an instrument. & A violin is an instrument. &\textbf{Logical Propositions:}\newline A: a violin (from Explanatory Sentence 1)\newline B: an instrument (from Explanatory Sentence 1)\newline \textbf{Logical Relations:}\newline Implies(A, B) \newline Implies(a violin, an instrument)\newline \textbf{Derived Implications:} none & \centering 0\arraybackslash & Valid \\\\
\bottomrule
\end{tabular}
\caption{An example of a verified logically valid explanation in the e-SNLI dataset without refinement.}
\label{tab:e-snli_example_0_table}
\end{table*}

\begin{table*}[t]
\centering
\small
\begin{tabular}{@{}p{1cm}p{3.5cm}p{3cm}p{4cm}p{1cm}p{1cm}@{}}  
\toprule
\textbf{Dataset} & \textbf{Problem} & \textbf{Explanation} & \textbf{Logic Info} & \textbf{Iteration} & \textbf{Validity} \\
\midrule
QASC & \textbf{Hypothesis}: Some viruses have a coating of phospholipids. & Some viruses have an envelope of phospholipids and proteins.\newline Proteins are sometimes coats of a virus. &\textbf{Logical Propositions:}\newline A: some viruses have an envelope of phospholipids and proteins (from Explanatory Sentence 1)\newline B: proteins are coats of a virus (from Explanatory Sentence 2)\newline \textbf{Logical Relations:}\newline none \newline \textbf{Derived Implications:} none & \centering 0\arraybackslash & Invalid \\\\
QASC & \textbf{Hypothesis}: Some viruses have a coating of phospholipids. & Some viruses have an envelope of phospholipids and proteins. \newline Proteins are sometimes coats of a virus. \newline An envelope can be considered a type of coating. \newline Phospholipids are a component of the envelope of some viruses. &\textbf{Logical Propositions:}\newline A: viruses have an envelope of phospholipids and proteins (from Explanatory Sentence 1)\newline B: proteins are coats of a virus (from Explanatory Sentence 2)\newline C: an envelope is a type of coating (from Explanatory Sentence 3)\newline D: phospholipids are a component of the envelope (from Explanatory Sentence 4) \newline \textbf{Logical Relations:}\newline Implies(A, D)\newline Implies(viruses have an envelope of phospholipids and proteins, phospholipids are a component of the envelope) \newline \textbf{Derived Implications:} none & \centering 1\arraybackslash & Valid \\\\
\bottomrule
\end{tabular}
\caption{An example of how an explanation in the QASC dataset can be refined in one iteration..}
\label{tab:qasc_example_1_table}
\end{table*}

\begin{table*}[t]
\centering
\small
\begin{tabular}{@{}p{1cm}p{3.5cm}p{3cm}p{4cm}p{1cm}p{1cm}@{}}  
\toprule
\textbf{Dataset} & \textbf{Problem} & \textbf{Explanation} & \textbf{Logic Info} & \textbf{Iteration} & \textbf{Validity} \\
\midrule
WorldTree & \textbf{Hypothesis}: A forest fire would cause deer to die or leave a woodland. & wildfire is when a forest catches fire.\newline fire causes harm to trees; to forests; to living things.\newline a deer is a kind of animal.\newline an animal is a kind of living thing.\newline killing means harming something causing death.\newline a deer lives in a forest.\newline woodland means forest.\newline natural disasters can cause animals to leave an environment.\newline a wildfire is a kind of natural disaster.\newline a forest is a kind of environment. &\textbf{Logical Propositions:}\newline A: wildfire (from Explanatory Sentence 1)\newline B: forest catches fire (from Explanatory Sentence 1)\newline C: fire causes harm to trees (from Explanatory Sentence 2)\newline D: fire causes harm to forests (from Explanatory Sentence 2)\newline E: fire causes harm to living things (from Explanatory Sentence 2)\newline F: deer (from Explanatory Sentence 3)\newline ...\newline Q: forest is a kind of environment (from Explanatory Sentence 10)\newline \textbf{Logical Relations:}\newline Equivalent(A, B) \newline Implies(C, D) \newline ...\newline Implies(M, Q) \newline \textbf{Derived Implications:} Implies(Not(O), Not(A)) \newline Implies(B, A)\newline ... \newline  Implies(L, Q) & \centering 0\arraybackslash & Invalid \\\\
WorldTree & \textbf{Hypothesis}: A forest fire would cause deer to die or leave a woodland. & Woodland means forest.\newline A wildfire is a kind of natural disaster.\newline A deer is a kind of animal.\newline An animal is a kind of living thing.\newline A forest is a kind of environment.\newline Natural disasters can cause animals to leave an environment.\newline Fire causes harm to trees; to forests; to living things.\newline A forest fire is a kind of wildfire.\newline A deer lives in a forest. &\textbf{Logical Propositions:}\newline A: woodland (from Explanatory Sentence 1)\newline B: forest (from Explanatory Sentence 1) ...\newline L: lives in a forest (from Explanatory Sentence 9)\newline \textbf{Logical Relations:}\newline Equivalent(A, B) \newline Implies(C, D) \newline ... \newline  Implies(I, Implies(B, Not(G))) \newline \textbf{Derived Implications:} Implies(E, Not(D)) \newline  Implies(E, Not(K)) \newline ...\newline Implies(Not(G), Not(E)) & \centering1\arraybackslash & Valid \\\\
\bottomrule
\end{tabular}
\caption{An example of how an explanation in the WorldTree dataset can be refined in one iteration.}
\label{tab:worldtree_example_1_table}
\end{table*}

\section{Prompts}
\label{appendix:prompt}
Tables \ref{tab:prompt_syntactic_parsing}, \ref{tab:prompt_refine_q}, \ref{tab:prompt_extracting}, and \ref{tab:prompt_proof} show the prompts we used for syntactic parsing, logical proposition extraction, logical relation extraction, and proof construction. Complete prompts details can be found at \href{https://github.com/neuro-symbolic-ai/faithful_and_robust_nli_refinement}{https://github.com/neuro-symbolic-ai/faithful\_and\_robust\_nli\_refinement}.

\begin{table*}[!ht]
\footnotesize
\setlength{\parindent}{0pt}  
\parbox{\textwidth}{  
\caption{Prompts used for syntactic parsing.}
\label{tab:prompt_syntactic_parsing}
}
\begin{tcolorbox}[width=\textwidth, 
    before skip=0pt,   
    after skip=0pt,   
    top=3pt,          
    bottom=3pt,
    left=3pt,
    right=3pt]

SYSTEM: You are an expert in linguistics. You will be provided with some sentences, please do a syntactic parse for each word in that sentence.
Some instructions:\\
1. You must give me the answer for all provided sentences.
2. Do not add any notes.
3. If no premise sentence provided, include it in the answer as none. 
4. Retain the answer words in their original form within the provided sentence.\\
USER: Here are some examples:\\
Hypothesis Sentence: \\
1. A woman is playing an instrument.\\
Subject: A woman\\
Verb Phrase: is playing an instrument\\
 - Main Verb: playing\\
 - Auxiliary Verb: is\\
Direct Object: an instrument\\
...\\
$<<<<<<<<<<<<<$\\
Provided sentences:\\
\\
Answer:
\end{tcolorbox}
\end{table*}

\begin{table*}[!ht]
\footnotesize
\setlength{\parindent}{0pt}  
\parbox{\textwidth}{  
\caption{Prompts used for extracting logical propositions and relations}
\label{tab:prompt_extracting}
}
\begin{tcolorbox}[width=\textwidth, 
    before skip=0pt,   
    after skip=0pt,   
    top=3pt,          
    bottom=3pt,
    left=3pt,
    right=3pt]

SYSTEM: You are an expert in symbolic reasoning. You will be provided with an explanation. You need to extract the logical propositions and the corresponding logical relations from the explanation.\\
USER: Here are some examples:\\
Provided Explanatory Sentences:\\
Explanatory Sentence 1: If it is raining, the grass will be wet.  \\
Explanatory Sentence 2: Having a picnic is equivalent to having a meal on the grass.\\
\\
Answer:
\\
Logical Propositions: \\
A: it is raining (from Explanatory Sentence 1)\\
B: the grass will be wet (from Explanatory Sentence 1)\\
C: having a picnic (from Explanatory Sentence 2)\\
D: having a meal on the grass (from Explanatory Sentence 2)\\
\\
Logical Relations:\\
Implies(A, B): $A \rightarrow B$\\
Equivalent(C, D): $C \leftrightarrow D$\\
\\
$<<<<<<<<<<<<<$\\
Provided Explanatory Sentences:\\
\\
Answer:\\
\\
Logical Propositions: \\
\\
Logical Relations:\\
\\
\end{tcolorbox}
\end{table*}

\begin{table*}[!ht]
\footnotesize
\setlength{\parindent}{0pt}  
\parbox{\textwidth}{  
\caption{Prompts used for refining quantifiers.}
\label{tab:prompt_refine_q}
}
\begin{tcolorbox}[width=\textwidth, 
    before skip=0pt,   
    after skip=0pt,   
    top=3pt,          
    bottom=3pt,
    left=3pt,
    right=3pt]
SYSTEM: You are an expert in semantics, formal language and neo-davidsonian event semantics. You will be provided with some sentences. 
These sentences have been transferred into Isabelle/HOL symbolic language. However, the quantifiers in the logical form may not be defined correctly.There might be missing variables after the quantifiers for arguments inside the parentheses of the predicate-argument forms of an axiom or a theorem. The quantifier may not reflect to real-world knowledge. Refine the logical forms if there are any quantifiers that are not defined correctly.\\
\\
Here are some examples:\\
Provided Iabelle code:\\
(* Explanation 1: Many consumers feed at more than one trophic level. *)\\
axiomatization where\\
  explanation\_1: "$\forall$x e. Consumer x $\longrightarrow$ (Feed e $\wedge$ Agent e x $\wedge$ At e y $\wedge$ $\mathit{MoreThanOneTrophicLevel}$ y)"\\
\\
Answer:\\
Explanation 1 states "Many consumers" and in real-world knowledge, some consumers are omnivores or generalists that feed across multiple trophic levels, but it use the universal quantifier '$\forall$' in explanation\_1. We should use the existential quantifier '$\exists$' instead. \\
For the quantifier variables in explanation\_1, the variable 'y' is missing.\\
...\\
$<<<<<<<<<<<<<$\\
Strictly follow my instructions. \\
\\
Provided Isabelle code:\\
\\
Answer:
\end{tcolorbox}
\end{table*}

\begin{table*}[!ht]
\footnotesize
\setlength{\parindent}{0pt}  
\parbox{\textwidth}{  
\caption{Prompts used for building proofs.}
\label{tab:prompt_proof}
}
\begin{tcolorbox}[width=\textwidth, 
    before skip=0pt,   
    after skip=0pt,   
    top=3pt,          
    bottom=3pt,
    left=3pt,
    right=3pt]

SYSTEM: You are an expert in Isabelle theorem prover, first-order logic and Davidsonian event semantics. You will be provided with premises, explanations and hypothesis sentences. You will be provided with an Isabelle code which consists of some axioms, a theorem hypothesis that needs to be proven. The logical form of axioms indicates some explanation sentences, the logical form after "assume asm:" indicates a premise sentence and the logical form after "shows" indicates a hypothesis sentence. The natural language form is stated as the comments. You will be provided with some logical propositions, logical relations and derived logical rules from the explanation sentences to help you construct the proof. You need to construct a proof about how to prove the theorem hypothesis in "proof -" and "qed" sections using the premise (logical form after "assume asm:") and explanations (axioms). The proof should be derived from the premise and explanation sentences. You don't need to state the automated theorem prover you will need to use. You just need to write a proof sketch.\\
Some instructions:\\
1. 'sorry' and ‘fix’ command is not allowed. \\
...\\
5. leave the automated theorem prover and proof tactic as <ATP>\\
...\\
$<<<<<<<<<<<<<$\\
Strictly follow my instructions. \\

Premise Sentence: \\
\\
Explanation Sentences: \\
\\
Hypothesis Sentence: \\
\\
Provided Isabelle Code:\\
\\
Logical Information: \\
\\
Known Information:\\
\\
Try to prove:\\
\\
Answer:\\
\end{tcolorbox}
\end{table*}

\end{document}